\documentclass[sigconf,nonacm,balance=false]{acmart}

\setcopyright{none}
\graphicspath{{figures/}}

\usepackage[utf8]{inputenc}
\usepackage{booktabs}    
\usepackage{multirow}    
\usepackage{amsmath}     
\usepackage{graphicx}    
\usepackage{tabularx}
\usepackage{array}

\AtBeginDocument{%
  }
\begin{document}
\title{OpenDepth: Repurposing Image Editing Priors for Robust Open-World Monocular Depth Estimation}
\author{Mengke Zhang}
\email{zmk1460997475@bupt.edu.cn}
\orcid{0009-0004-1250-5505}
\affiliation{%
  \institution{International School, Beijing University of Posts and Telecommunications}
  \city{Beijing}
  \country{China}
}

\author{Tianyi Yang}
\email{3662748738@bupt.edu.cn}
\orcid{0009-0006-5709-330X}
\affiliation{%
  \institution{International School, Beijing University of Posts and Telecommunications}
  \city{Beijing}
  \country{China}
}

\author{Xinzhuo Li}
\email{xinzhuo.li@se23.qmul.ac.uk}
\orcid{0009-0002-7898-9755}
\affiliation{%
  \institution{International School, Queen Mary University of London}
  \city{Beijing}
  \country{China}
}

\author{Anlong Ming}
\authornote{Corresponding author.}
\email{mal@bupt.edu.cn}
\orcid{0000-0003-2952-7757}
\affiliation{%
  \institution{School of Computer Science, Beijing University of Posts and Telecommunications}
  \city{Beijing}
  \country{China}
}
\renewcommand{\shortauthors}{Mengke Zhang, Tianyi Yang, Xinzhuo Li, and Anlong Ming}
\begin{abstract}
Recently, the integration of generative priors has revolutionized Monocular Depth Estimation (MDE), achieving state-of-the-art (SOTA) performance with remarkably high data efficiency. However, although methods based on Text-to-Image (T2I) models excel in general scenes, their efficacy degrades sharply when confronted with open-world environmental perturbations, such as illumination variations and adverse weather. Since T2I priors are optimized for semantic synthesis rather than preserving pixel-level structure, they struggle to decouple geometry from appearance under severe visual shifts, easily triggering structural hallucinations. Therefore, our core insight is to model open-world depth estimation as consistent geometric inference across multi-style images, fully leveraging the powerful structure-preserving editing priors of Image-to-Image (I2I) models. Based on this, we propose OpenDepth, a robust open-world MDE framework that introduces ``Intra-Domain Awareness''—the capability to map diverse stylistic inputs sharing identical geometry into a unified latent space. To achieve this, we first align shallow features within the Diffusion Transformer (DiT) to anchor degraded observations to unified geometric representations. Additionally, we introduce a joint RGB latent reconstruction task, requiring the model to recover clean RGB information from degraded inputs. This multi-stage strategy effectively decouples environmental style interference, compelling the model to focus strictly on underlying geometric recovery. Extensive evaluations show that OpenDepth sets new SOTA records for robust zero-shot depth estimation across standard and extreme weather benchmarks, demonstrating exceptional structural integrity and adaptability.
\par
\noindent Code: \url{https://github.com/zmkshakespar/OpenDepth}.
\end{abstract}

\maketitle

\section{Introduction}
Monocular Depth Estimation \cite{eigen2014depth, ming2021deep, bhat2021adabins, yang2024depth} serves as a cornerstone for a wide range of downstream applications, including autonomous driving \cite{wang2019pseudo}, robotic navigation \cite{tai2018socially}, and augmented reality \cite{du2020depthlab}. Despite its undeniable importance, achieving high-fidelity depth maps remains a formidable challenge due to the task's inherent ill-posed nature and the scarcity of large-scale, precisely labeled datasets. Recently, the field has witnessed a paradigm shift with the integration of generative priors. Pre-trained Text-to-Image diffusion models, such as Stable Diffusion \cite{rombach2022high}, have demonstrated a remarkable capacity to capture complex scene statistics. By treating depth as a specialized image generation modality, these models have achieved SOTA performance in zero-shot geometric estimation.

\begin{figure}[t]
  \centering
  \includegraphics[width=\linewidth]{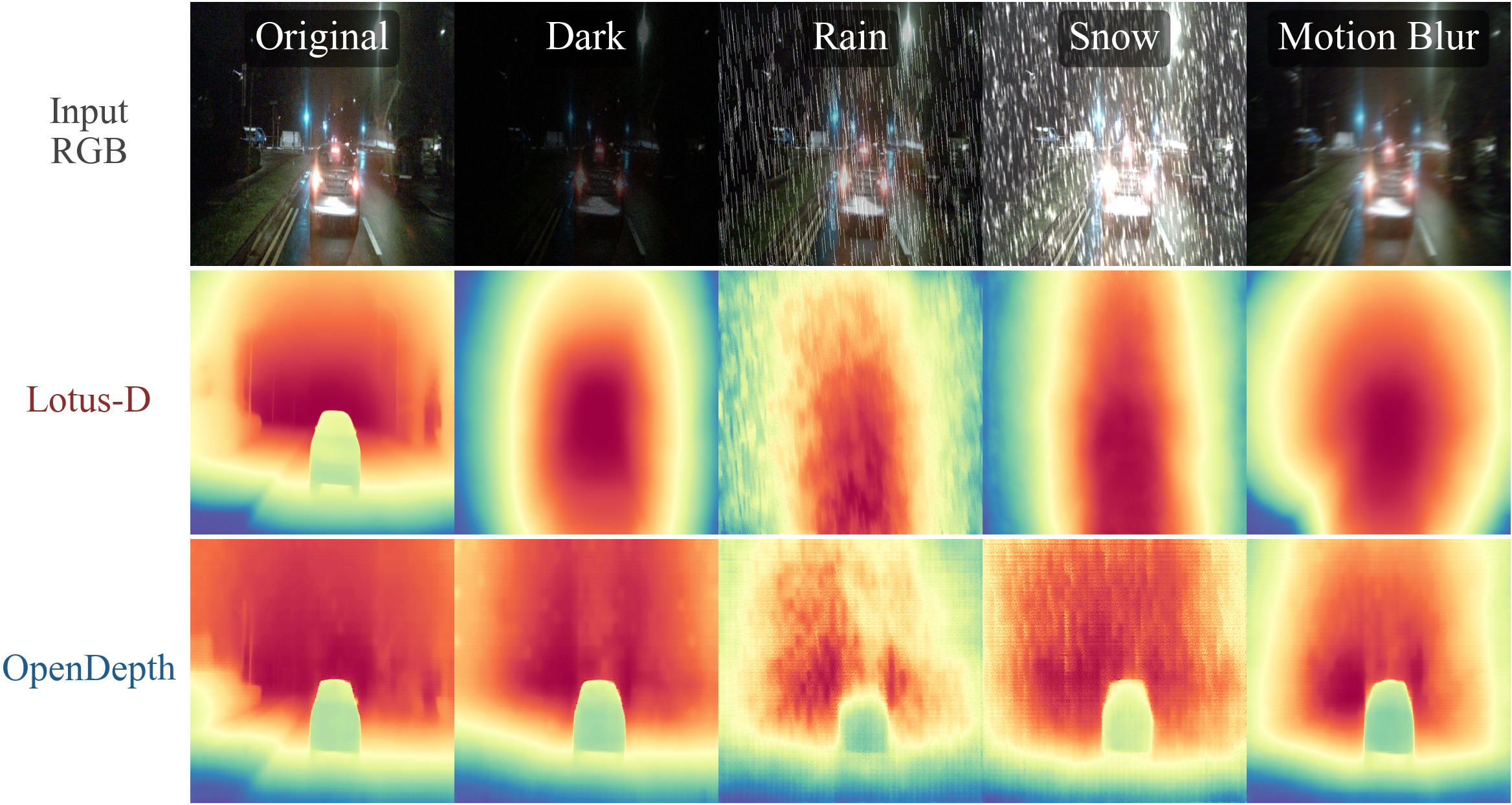}
\caption{\textbf{Standalone comparison of T2I and I2I priors under perturbations (zero-shot).} In a separate controlled experiment, both models are trained on the same 59K samples from Hypersim \cite{roberts2021hypersim} and Virtual KITTI \cite{cabon2020virtual}. The T2I model (Lotus-D \cite{he2024lotus}) suffers from severe structural hallucinations under visual degradations, while the I2I baseline better preserves scene geometry, suggesting stronger inherent robustness for depth estimation.}
  \label{fig:motivation}
\end{figure}

However, existing models struggle with complex environmental perturbations (e.g., adverse weather, motion blur) in open-world tasks. Directly fine-tuning them on degraded samples remains sub-optimal due to the scarcity of comprehensively annotated real-world data. While recent methods leverage generative priors to achieve SOTA performance with limited data, relying on T2I models introduces new bottlenecks. Optimized primarily for semantic synthesis, T2I priors frequently produce severe structural hallucinations under visual degradations. Fundamentally, monocular depth estimation is an Image-to-Image translation task, making T2I priors intrinsically misaligned with its strict geometric objectives. To isolate this, we conduct a matched-data comparison (Figure \ref{fig:motivation}) training both a T2I model (Lotus-D \cite{he2024lotus}) and our I2I baseline on the identical 59k samples. Under varying perturbations, the I2I baseline successfully preserves scene geometry, contrasting sharply with the T2I model's structural failures. This highlights the superior inherent robustness of I2I priors for geometric tasks, motivating our I2I-based foundation.

To overcome this fundamental limitation, our core insight is to reframe open-world depth estimation as a unified geometric inference process across diverse visual styles. In this context, Image-to-Image editing models naturally align with the task, as they excel at accommodating various stylistic transformations while strictly preserving the underlying structural integrity. Nevertheless, the mere injection of editing priors is insufficient to fully resolve the challenge. Under extreme environmental degradations, severe input distribution shifts still trigger alignment deviations within the feature space, hindering simple, direct depth regression from extracting pure geometric cues. Consequently, we argue that a robust depth estimator must possess ``Intra-Domain Awareness''—the capability to actively map and anchor observations from diverse degradation conditions into a unified, clean latent domain prior to executing core geometric reasoning. This mechanism is indispensable for truly unleashing the geometric perception potential of editing priors in complex open-world environments.

Driven by this philosophy, we introduce \textbf{OpenDepth}, a robust open-world MDE framework. Moving beyond existing approaches that focus exclusively on regressing depth from degraded features, our method prioritizes global alignment within the feature space. Specifically, we implement a multi-stage alignment and reconstruction mechanism: First, we align image features from the early Single Stream blocks to establish robust correspondence between perturbed observations and their unified geometric structures. Subsequently, by leveraging the inherent bipartite structure of the DiT output sequence, we introduce unified RGB latent reconstruction as a joint task alongside depth estimation. This joint optimization compels the model to reconstruct unperturbed RGB information from diverse conditional inputs, thereby ensuring the latent features conform to the unified latent domain established by the editing prior. This strategy effectively enables the model to decouple environmental style interference and focus exclusively on recovering high-fidelity geometry.

The contributions of this work are threefold: 
$\bullet$ \textbf{Systematic Analysis of Generative Priors:} We provide a comprehensive analysis of the limitations of T2I-based generative priors in MDE, revealing that their stochastic nature and lack of structural constraints lead to performance degradation in complex environments. We demonstrate the superiority of shifting toward I2I editing models as the foundation for robust geometric tasks. 
$\bullet$ \textbf{Unified Latent Alignment Framework:} We propose an innovative alignment mechanism that transcends direct regression by operating in a constrained latent space. By integrating unified RGB reconstruction and initial DiT block alignment, we ensure that diverse visual observations are projected into a unified latent domain, establishing a robust ``many-to-one'' mapping from perturbed inputs to ground-truth depth. 
$\bullet$ \textbf{SOTA Performance \& High Data Efficiency:} OpenDepth sets new SOTA records for robust zero-shot depth estimation across standard and degraded benchmarks. Notably, it achieves exceptional detail preservation in complex open-world scenes using only 79K training samples, demonstrating remarkable data efficiency.

\begin{figure*}[t]
    \centering
    \includegraphics[width=0.80\textwidth]{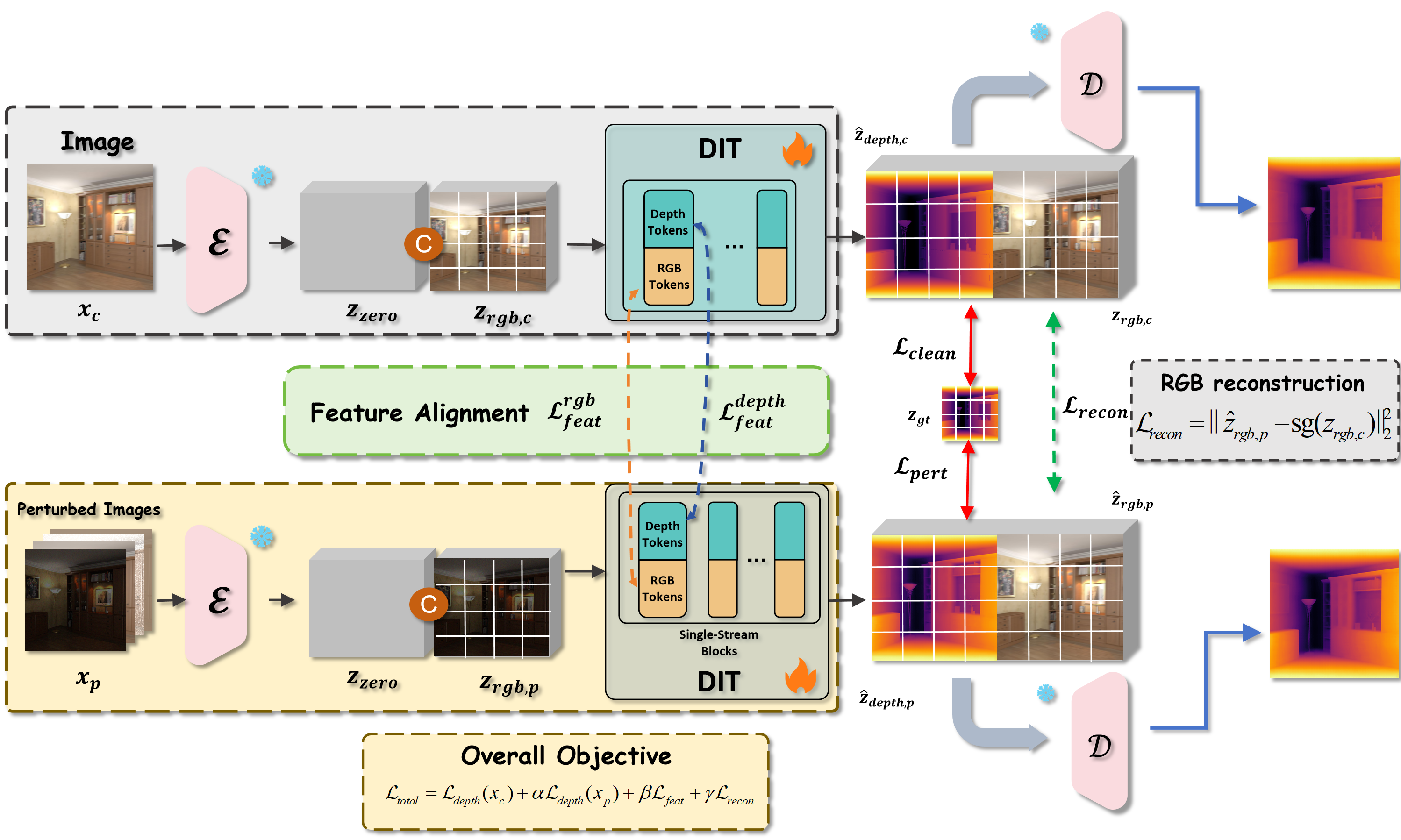}
   \caption{\textbf{Overview of the OpenDepth framework.} 
    To harness editing priors for robust open-world depth estimation, we introduce \textbf{Latent Feature Alignment} ($\mathcal{L}_{feat}$) and \textbf{Unified RGB Reconstruction} ($\mathcal{L}_{recon}$). 
    This dual-constraint strategy projects diverse degraded inputs into a unified latent domain, strictly decoupling environmental styles from geometric structures.}
    \label{fig:pipeline}
\end{figure*}

\section{Related Work}

\subsection{Foundation MDE Model}

MDE has rapidly evolved from early supervised and self-supervised methods \cite{eigen2014depth, godard2019digging, zhou2017unsupervised, ming2021deep} into powerful foundation models with formidable zero-shot generalization. Discriminative approaches established robust multi-source training paradigms \cite{ranftl2020towards, ranftl2021vision, yin2020diversedepth, yin2021leres, eftekhar2021omnidata}, while subsequent works improved metric accuracy and structural fidelity via transformer decoding, internal discretization, and scale decomposition \cite{bhat2021adabins, yuan2022newcrfs, piccinelli2023idisc, yin2023metric3d, hu2024metric3dv2, bhat2023zoedepth, piccinelli2024unidepth, bochkovskii2024depth, li2024patchfusion, zhu2024scaledepth, wang2025moge, liu2024sm4depth}. 
The Depth Anything series \cite{yang2024depth, yang2024depthv2} pushed performance boundaries via large-scale semi-supervised learning, while its latest iteration~\cite{lin2025depth} further extended this paradigm to unified multi-view and monocular geometry recovery with a plain transformer architecture. Lightweight discriminative frameworks also proved that efficiency and accuracy can coexist \cite{ren2026anydepth}. Concurrently, generative models repurposed T2I diffusion priors for dense prediction \cite{zhao2023unleashing, ke2024repurposing, fu2024geowizard}, with subsequent works optimizing inference via Flow Matching and deterministic single-step paradigms \cite{gui2025depthfm, he2024lotus, xu2024matters, song2025depthmaster}, improving conditioning and detail refinement \cite{patni2024ecodepth, zhang2024betterdepth, garcia2025fine, yang2025ogdepth, liu2025tpdepth}, 
unifying generation and dense perception \cite{zheng2024diff}, extending to video sequences \cite{hu2025depthcrafter}, and adapting diffusion priors for self-supervised learning \cite{wang2025jasmine}. However, as argued by FE2E \cite{wang2025editor}, T2I models exhibit inherent stochasticity and lack pixel-level structural constraints, motivating a shift toward image editing models \cite{brooks2023instructpix2pix, liu2025step1x, wu2025qwen} whose intrinsic ``editing priors'' naturally maintain structural integrity. This has given rise to a new class of I2I-based dense geometry estimators, represented by FE2E \cite{wang2025editor} and Edit2Perceive \cite{shi2025edit2perceive}, which demonstrate that repurposing image editing models yields superior structural fidelity over their T2I counterparts for depth and normal estimation.

Compared to existing I2I schemes (e.g., FE2E \cite{wang2025editor}, Edit2Perceive \cite{shi2025edit2perceive}), OpenDepth addresses a critical gap: while existing methods leverage editing models to regress depth directly, severe input shifts in extreme environments still cause feature space misalignments. Rather than simply migrating editing features, OpenDepth introduces an ``Intra-Domain Awareness'' alignment strategy. By incorporating a unified RGB reconstruction task and shallow latent space alignment at the earliest feature extraction stages, we compel the model to project diverse degraded observations into a unified latent domain defined by the editing prior, enabling superior consistency and noise resistance.

\subsection{Robust Monocular Depth Estimation}

Achieving robustness under open-world perturbations is critical, as systematic benchmarks \cite{hendrycks2019benchmarking, kar20223d, kong2023robodepth} reveal severe degradation in SOTA models under adverse weather and illumination. Early domain adaptation bridged the synthetic-to-real gap via adversarial alignment or image translation \cite{godard2017unsupervised, vankadari2020unsupervised, wang2021regularizing, gasperini2023robust}. Subsequently, self-supervised methods tackled photometric consistency violations through joint nighttime enhancement \cite{zheng2023steps}, frequency-aware regularization \cite{chen2023frequency}, masked image modeling \cite{chawla2023image}, curriculum contrastive learning \cite{wang2024weatherdepth, saunders2023self}, depth consistency regularization \cite{song2023ec}, dynamic video training \cite{sun2023sc}, IMU-guided learning \cite{zhang2022towards}, dual-branch all-day estimation \cite{zhang2024glocalfuse}, distribution compensation \cite{yang2024self}, foundation-model-assisted stereo \cite{vankadari2024dusk}, stabilized adversarial generalization \cite{yao2024improving}, and motion-structure priors \cite{yan2025synthetic}. In the foundation model era, parameter-efficient adaptation has emerged as the dominant strategy for weather robustness via knowledge distillation \cite{ zeng2025depthdark} 
and LoRA variants \cite{yang2024multi, yan2025er}. Generative approaches also explore clean-image synthesis \cite{jiang2025always} or joint dehazing \cite{zhang2024depth}. Despite this progress, current state-of-the-art open-world models, such as DepthAnything-AC \cite{sun2025depth}, primarily rely on unsupervised output-consistency and spatial distance constraints. 

Diverging from current SOTA methods like DepthAnything-AC \cite{sun2025depth}, which primarily rely on discriminative architectures and massive data-level augmentations, OpenDepth pioneers the integration of generative Image-to-Image (I2I) editing priors for robust open-world perception. By leveraging the inherent structural preservation capabilities of these foundation models, our approach achieves superior robustness with remarkably high data efficiency, utilizing merely 79K training samples. Mechanistically, rather than enforcing complex output-level consistency, we propose a concise ``Latent Alignment'' pathway. By projecting diverse degraded inputs into a unified, style-invariant latent domain, this mechanism fundamentally eliminates environmental interference at the feature source, ensuring exceptional structural integrity in complex scenarios.

\section{Methodology}
In this section, we introduce \textbf{OpenDepth}, a robust Monocular Depth Estimation framework built upon the STEP1X-Edit foundation model (overview shown in Figure~\ref{fig:pipeline}). We first review the Multi-Modal Diffusion Transformer (MMDiT) preliminaries and analyze the geometric advantages of I2I editing priors. Finally, we detail our core contributions—Latent Feature Alignment and Unified RGB Reconstruction—which construct a unified latent space for robust geometric perception.

\subsection{Preliminaries: Multi-Modal Diffusion Transformer}
Unlike conventional U-Nets, the MMDiT backbone in OpenDepth processes serialized patch embeddings to handle RGB conditions, depth targets, and text prompts. 

Specifically, the input RGB image $x$ and a target depth $d$ are encoded via a pre-trained VAE into latents $z_{rgb}$ and $z_{depth}$, which are patchified and concatenated into a joint spatial input $\mathcal{T}_{in}$. The MMDiT processes $\mathcal{T}_{in}$ alongside the text sequence $\mathcal{T}_{txt}$ through two stages:

\textbf{1. Double Stream Blocks:} The spatial and text sequences maintain independent representations but interact via shared attention:
\begin{equation}
    \mathcal{T}_{img}^{(k)}, \mathcal{T}_{txt}^{(k)} = \text{DoubleStreamBlock}(\mathcal{T}_{img}^{(k-1)}, \mathcal{T}_{txt}^{(k-1)}, \mathcal{E}_{time})
\end{equation}
where $\mathcal{E}_{time}$ denotes the timestep embedding, and $\mathcal{T}_{img}^{(0)} = \mathcal{T}_{in}$.

\textbf{2. Single Stream Blocks:} After $K$ double stream blocks, the sequences are concatenated for deep semantic fusion:
\begin{equation}
    \mathcal{T}_{joint}^{(l)} = \text{SingleStreamBlock}(\mathcal{T}_{joint}^{(l-1)}, \mathcal{E}_{time})
\end{equation}
where $\mathcal{T}_{joint}^{(0)} = [\mathcal{T}_{txt}^{(K)} \parallel \mathcal{T}_{img}^{(K)}]$. Finally, the output sequence is un-patchified into the predicted depth latent $\hat{z}_{depth}$ and the reconstructed RGB latent $\hat{z}_{rgb}$.

To clarify the diffusion formulation, OpenDepth operates in a deterministic, discriminative mode. Instead of iterative denoising, the target depth $d$ is strictly zero-initialized, and the MMDiT performs a single-step direct regression at a fixed timestep ($t=1.0$). Consequently, all training objectives—including $\mathcal{L}_{feat}$ and $\mathcal{L}_{recon}$—are computed exclusively at this fixed timestep.

\subsection{Advantages of I2I Editing Priors: A Motivational Analysis}

Existing generative MDE methods primarily rely on T2I models (e.g., Stable Diffusion \cite{rombach2022high}). However, T2I models are optimized for semantic synthesis driven by text prompts and lack explicit constraints to maintain pixel-level structural integrity. When input images are corrupted by environmental perturbations (e.g., rain, haze, low light, or noise), T2I priors tend to produce structural hallucinations or blurry boundaries, as the degraded visual features deviate from their expected distribution. Conversely, I2I editing models possess inherent transformation priors: they are trained to alter specific attributes (style or domain) while strictly preserving the underlying geometry of the source image.

\begin{figure}[t]
    \centering
    \includegraphics[width=\linewidth]{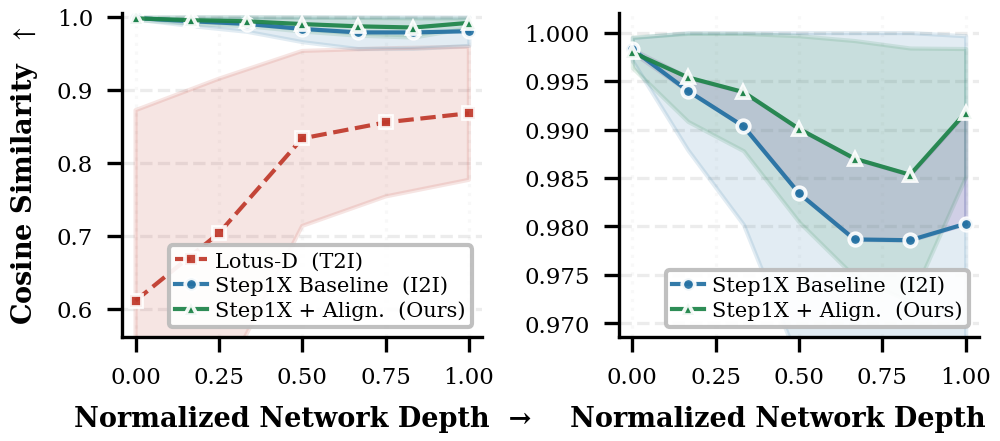}
    \caption{Cosine similarity analysis in the feature space across normalized network depths. (Left) Comparison between T2I (Lotus-D) and I2I (OpenDepth) priors. The I2I baseline exhibits significantly higher and more stable similarity than the T2I model, demonstrating inherent intra-domain awareness. (Right) Effect of our Latent Feature Alignment.}
    \label{fig:cosine_similarity}

\end{figure}

\textbf{Motivational Experiment and Analysis.} To validate this hypothesis, we evaluate a representative T2I-based depth model (Lotus-D \cite{he2024lotus}) against our I2I baseline (without robustness fine-tuning) by calculating the cosine similarity of intermediate spatial features between clean and perturbed images across different layers. As shown in Figure~\ref{fig:cosine_similarity}, the T2I model struggles to recognize the geometric essence, exhibiting extremely low similarity in initial layers. Conversely, the I2I model penetrates the perturbations early on, capturing geometric features highly consistent with the clean input and maintaining this high similarity throughout deeper layers. This phenomenon demonstrates that I2I models possess an inherent ``intra-domain awareness'' that resists visual degradation, serving as the logical cornerstone for adopting an I2I editing model as our foundation for robust MDE.

\subsection{Latent Feature Alignment}
Although I2I priors are inherently structure-preserving, severe perturbations can still cause feature drift after cross-modal fusion. We therefore apply Latent Feature Alignment to image tokens in the first $L$ Single Stream blocks, where features already capture preliminary cross-modal context while remaining close to the input structure.

Our goal is to establish a rigid mapping ensuring that a perturbed image $x_p$ yields identical, fine-grained geometric features to its clean version $x_c$. As introduced in Section 3.1, during the Single Stream Block phase, the network has completed preliminary cross-modal interactions and begins extracting deep spatial features. Let $\mathcal{T}_{joint}^{(l)} \in \mathbb{R}^{B \times (N_{txt} + N_{img}) \times C}$ denote the output sequence of the $l$-th Single Stream Block. We first slice the sequence to isolate the spatial tokens corresponding to the image and depth:

\begin{equation}
    F^{(l)} = \mathcal{T}_{joint}^{(l)}[:, N_{txt}:, :] \in \mathbb{R}^{B \times N_{img} \times C}
\end{equation}

For each sample $b$ in the batch, we flatten its spatial features into a vector $f_b^{(l)} \in \mathbb{R}^{N_{img} \cdot C}$ and compute a per-sample cosine similarity loss. We enforce this feature consistency in the first $L$ shallow Single Stream Blocks:

\begin{equation}
    \mathcal{L}_{feat} = \frac{1}{L} \sum_{l=1}^{L} \frac{1}{B} \sum_{b=1}^{B} \left[ 1 - \frac{\text{sg}(f_{c,b}^{(l)}) \cdot f_{p,b}^{(l)}}{\|\text{sg}(f_{c,b}^{(l)})\|_2 \|f_{p,b}^{(l)}\|_2} \right]
\end{equation}

where $\text{sg}(\cdot)$ denotes the stop-gradient operation. Similar to the design in Section~3.4, the clean features serve as immutable anchors, ensuring that only the perturbed branch is updated to align with the canonical geometric representation.

By aligning features in the early Single Stream blocks (we use $L=4$ in practice), the network learns to suppress environmental noise early and extract cleaner depth-relevant features. As shown in Figure~\ref{fig:cosine_similarity}, this alignment improves feature similarity throughout the network, especially in deeper layers, yielding more consistent geometric representations. Ablations on different alignment depths are provided in the Appendix.

\subsection{Unified RGB Reconstruction in Latent Space}

\begin{figure}[t]
  \centering
  \includegraphics[width=\linewidth]{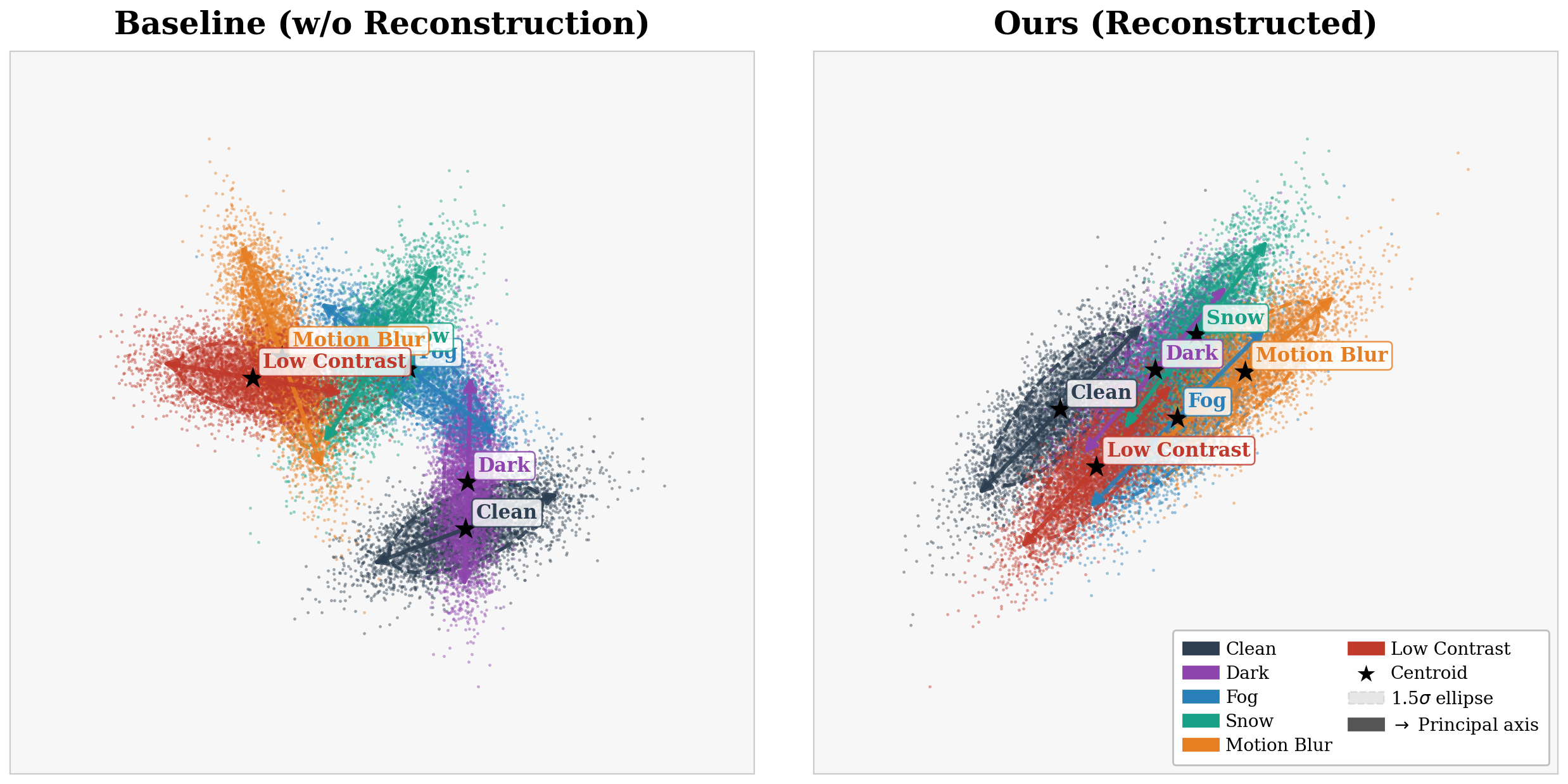}
  \caption{\textbf{t-SNE visualization of deep features across diverse environmental styles.} 
  (Left) The \textbf{Baseline} exhibits scattered feature clusters dominated by environmental styles. 
  (Right) Our \textbf{RGB Reconstruction} perfectly aligns these diverse perturbation clusters into a \textbf{unified latent domain}, demonstrating true style-invariance for consistent geometric perception.}
  \label{fig:tsne}
\end{figure}

While feature alignment ensures the purity of shallow features, we further expect the model to ``see through'' appearance variations to reach a unified semantic understanding of scenes with identical geometry. To this end, we introduce an implicit restoration task: Unified RGB Reconstruction in the latent space.
\begin{table*}[t]
\centering
\caption{\textbf{Zero-shot relative depth estimation results on general benchmarks.} Comparison with general depth estimation models and robust depth estimation models. Best results are in \textbf{bold}, and second best are \underline{underlined}. * indicates models based on pre-trained Stable Diffusion.}
\label{tab:general_benchmarks}
\resizebox{0.9\textwidth}{!}{
\begin{tabular}{l c| c c| c c| c c| c c | c c | c}
\toprule
\multirow{2}{*}{\textbf{Method}} & \multirow{2}{*}{\textbf{Encoder}} & \multicolumn{2}{c}{\textbf{KITTI \cite{geiger2012we}}} & \multicolumn{2}{c}{\textbf{NYUv2 \cite{silberman2012indoor}}} & \multicolumn{2}{c}{\textbf{Sintel \cite{butler2012naturalistic}}} & \multicolumn{2}{c}{\textbf{ETH3D \cite{schops2017multi}}} & \multicolumn{2}{c|}{\textbf{DIODE(indoor) \cite{vasiljevic2019diode}}} & \multirow{2}{*}{Avg Rank$\downarrow$} \\
\cmidrule(lr){3-4} \cmidrule(lr){5-6} \cmidrule(lr){7-8} \cmidrule(lr){9-10} \cmidrule(lr){11-12}
& & AbsRel$\downarrow$ & $\delta1\uparrow$ & AbsRel$\downarrow$ & $\delta1\uparrow$ & AbsRel$\downarrow$ & $\delta1\uparrow$ & AbsRel$\downarrow$ & $\delta1\uparrow$ & AbsRel$\downarrow$ & $\delta1\uparrow$ & \\
\midrule
\multicolumn{13}{c}{\textbf{General Depth Estimation Model}} \\
\midrule
DepthPro \cite{bochkovskii2024depth} & ViT-S & 6.8 & 94.7 & 5.2 & 97.2 & 24.0 & 68.2 & 5.7 & 96.3 & \underline{6.0} & \underline{95.2} & 4.3 \\
DepthAnything V1 \cite{yang2024depth} & ViT-S & 8.2 & 93.6 & 5.3 & 97.2 & 21.1 & \underline{73.6} & 6.4 & 96.6 & 7.8 & 94.0 & 5.2 \\
DepthAnything V2 \cite{yang2024depthv2} & ViT-S & 8.3 & 93.4 & 5.1 & 97.3 & 24.0 & 71.2 & 6.4 & 96.9 & 7.3 & 94.3 & 4.9 \\
Marigold$_{v1.1}^*$ \cite{ke2024repurposing} & VAE & 9.9 & 91.6 & 5.5 & 96.4 & 42.5 & 53.5 & 6.5 & 95.9 & 31.0 & 87.5 & 10.2 \\
Lotus-D$^*$ \cite{he2024lotus} & VAE & 8.8 & 91.2 & 5.2 & 96.8 & 29.9 & 64.9 & 5.7 & 96.7 & 8.3 & 92.1 & 7.3 \\
Diffusion-E2E-FT$^*$ \cite{garcia2025fine} & VAE & 9.6 & 92.1 & 5.4 & 96.5 & 45.7 & 57.1 & 6.4 & 95.9 & 28.6 & 87.7 & 9.3 \\
FE2E \cite{wang2025editor} & VAE & \underline{6.6} & \underline{96.0} & \underline{4.1} & \underline{97.7} & \underline{21.0} & 73.1 & \underline{3.8} & \textbf{98.7} & 6.2 & 94.8 & 2.2 \\
\midrule
\multicolumn{13}{c}{\textbf{Robust Depth Estimation Model}} \\
\midrule
DynaDepth \cite{zhang2022towards} & ResNet & 8.5 & 91.6 & 21.2 & 63.0 & 45.8 & 39.1 & 19.2 & 70.3 & 18.0 & 73.7 & 12.1 \\
EC-Depth \cite{song2023ec} & ViT-S & 7.5 & 93.1 & 22.0 & 60.6 & 44.7 & 40.5 & 18.1 & 77.9 & 17.2 & 76.0 & 10.2 \\
STEPS \cite{zheng2023steps} & ResNet & 20.4 & 68.3 & 22.0 & 60.6 & 45.8 & 37.5 & 19.7 & 69.9 & 21.4 & 65.2 & 13.7 \\
robustdepth \cite{saunders2023self} & ViT-S & 7.6 & 92.9 & 17.4 & 71.8 & 41.1 & 43.6 & 18.5 & 77.6 & 18.2 & 74.1 & 10.2 \\
weather-depth \cite{wang2024weatherdepth} & ViT-S & 7.5 & 93.2 & 16.7 & 73.3 & 41.9 & 44.3 & 18.1 & 77.1 & 17.0 & 76.0 & 9.2 \\
DepthAnything-AC \cite{sun2025depth} & ViT-S & 8.3 & 93.4 & 5.1 & 97.3 & 23.5 & 71.5 & 6.5 & 96.8 & 7.3 & 94.1 & 5.1 \\
\midrule
\textbf{OpenDepth (Ours)} & VAE & \textbf{6.3} & \textbf{96.2} & \textbf{3.9} & \textbf{98.0} & \textbf{19.9} & \textbf{76.2} & \textbf{3.7} & \underline{98.4} & \textbf{5.5} & \textbf{95.3} & \textbf{1.1} \\
\bottomrule
\end{tabular}
}
\end{table*}

By leveraging the inherent bipartite structure of the MMDiT output sequence, the architecture naturally produces a reconstructed RGB latent $\hat{z}_{rgb}$ alongside the depth estimate. Specifically, for a perturbed input, the final spatial output sequence $\mathcal{T}_{img}^{(out)}$ is first split along the channel dimension into depth prediction tokens $\mathcal{T}_{depth, p}$ and RGB reconstruction tokens $\mathcal{T}_{rgb, p}$. Subsequently, through an inverse unpatchify operation, we restore $\mathcal{T}_{rgb, p}$ into a 2D latent feature map:
\begin{equation}
    \hat{z}_{rgb, p} = \text{Unpatchify}(\mathcal{T}_{rgb, p})
\end{equation}
Rather than discarding this output, we utilize it to unify the perturbed and clean branches at the feature level. When the network receives a perturbed latent $z_{rgb, p}$, we supervise its RGB output $\hat{z}_{rgb, p}$ using the clean RGB latent $z_{rgb, c}$ (extracted from the unperturbed image $x_c$ via the VAE). The cross-domain reconstruction loss is defined as:
\begin{equation}
    \mathcal{L}_{recon} = \| \hat{z}_{rgb,p} - \text{sg}(z_{rgb, c}) \|_2^2
\end{equation}
where $\text{sg}(\cdot)$ denotes the stop-gradient operation. The stop-gradient operation ensures that the clean RGB latent serves strictly as an immutable anchor. This prevents erroneous gradient flow from degrading the clean representations, forcing the perturbed branch to actively align with the canonical domain rather than allowing the two branches to collapse toward a sub-optimal middle state.

\textbf{Feature Clustering Analysis.} To verify that the reconstruction module forces the model to ignore visual styles and extract pure geometric cues, we perform t-SNE visualization (Figure~\ref{fig:tsne}) on deep features (MMDiT Single Stream Block 37) across 20 scenes under 6 distinct visual perturbations. Without the reconstruction constraint, feature clusters for various perturbation types scatter diversely across the latent space with misaligned extension directions, revealing that the \textbf{Baseline}'s internal representation remains heavily entangled with surface-level visual noise. By integrating the reconstruction mechanism, \textbf{OpenDepth} achieves perfectly aggregated feature clusters across all perturbation types with fully aligned principal distributional axes, demonstrating that our model effectively filters out appearance interference (e.g., fog, blur, or low-light) and extracts consistent underlying geometry, achieving true perturbation invariance.

\subsection{Overall Objective}
During training, each batch contains a clean image $x_c$ and its randomly perturbed counterpart $x_p$, sharing the same ground-truth depth $d_{gt}$. The depth regression loss $\mathcal{L}_{depth}$ computes the masked Mean Squared Error (MSE) between the predicted and ground-truth depth latents. The overall robust training objective is a weighted sum of the clean depth loss, perturbed depth loss, feature alignment loss, and cross-domain RGB reconstruction loss:
\begin{equation}
    \mathcal{L}_{total} = \mathcal{L}_{depth}(x_c) + \alpha \mathcal{L}_{depth}(x_p) + \beta \mathcal{L}_{feat} + \gamma \mathcal{L}_{recon}
\end{equation}

where $\alpha$, $\beta$, and $\gamma$ are hyperparameters that balance perturbation robustness, feature consistency, and the contribution of latent reconstruction. Empirically, we set $\alpha=1$, $\beta=1$, and $\gamma=1$ as the optimal configuration, with detailed sensitivity ablations provided in the Appendix. This rigorous, multi-stage alignment strategy effectively capitalizes on editing priors to complete dense geometric tasks in complex scenarios.
\begin{table*}[t]
\centering
\caption{\textbf{Zero-shot relative depth estimation results on open-world  benchmarks.} Comparison with general depth estimation models and robust depth estimation models.  * indicates models based on pre-trained Stable Diffusion.}
\label{tab:open_world}
\resizebox{0.9\textwidth}{!}{
\begin{tabular}{l c |c c |c c| c c| c c| c c | c}
\toprule
\multirow{2}{*}{\textbf{Method}} & \multirow{2}{*}{\textbf{Encoder}} & \multicolumn{2}{c}{\textbf{NuScenes-night \cite{caesar2020nuscenes}}} & \multicolumn{2}{c}{\textbf{RobotCar-night \cite{maddern20171}}} & \multicolumn{2}{c}{\textbf{DS-rain \cite{yang2019drivingstereo}}} & \multicolumn{2}{c}{\textbf{DS-cloud \cite{yang2019drivingstereo}}} & \multicolumn{2}{c|}{\textbf{DS-fog \cite{yang2019drivingstereo}}} & \multirow{2}{*}{Avg Rank$\downarrow$} \\
\cmidrule(lr){3-4} \cmidrule(lr){5-6} \cmidrule(lr){7-8} \cmidrule(lr){9-10} \cmidrule(lr){11-12}
& & AbsRel$\downarrow$ & $\delta1\uparrow$ & AbsRel$\downarrow$ & $\delta1\uparrow$ & AbsRel$\downarrow$ & $\delta1\uparrow$ & AbsRel$\downarrow$ & $\delta1\uparrow$ & AbsRel$\downarrow$ & $\delta1\uparrow$ & \\
\midrule
\multicolumn{13}{c}{\textbf{General Depth Estimation Model}} \\
\midrule
DepthPro \cite{bochkovskii2024depth} & ViT-S & 21.8 & 66.9 & 23.7 & 53.4 & 12.4 & \underline{84.1} & 15.8 & 77.9 & 10.2 & 89.2 & 5.1 \\
DepthAnything V1 \cite{yang2024depth} & ViT-S & 23.2 & 67.9 & 23.9 & 51.8 & 13.3 & 81.9 & 15.0 & 80.1 & \underline{9.8} & 89.1 & 5.6 \\
DepthAnything V2 \cite{yang2024depthv2} & ViT-S & 20.0 & 72.5 & 23.9 & 51.8 & 12.5 & 84.0 & 15.1 & 79.8 & 10.3 & 89.0 & 5.4 \\
MoGe\cite{wang2025moge} & ViT-L & 32.9 & 41.5 & 37.9 & 27.6 & 23.7 & 53.2 & 24.5 & 52.7 & 24.4 & 47.9 & 13.4 \\
Marigold$_{v1.1}^*$ \cite{ke2024repurposing} & VAE & 32.2 & 49.2 & 24.4 & 60.2 & 16.0 & 79.7 & 18.5 & 74.9 & 13.4 & 85.9 & 11.2 \\
Lotus-D$^*$ \cite{he2024lotus} & VAE & 22.1 & 66.8 & 24.4 & 53.4 & \underline{12.2} & 83.8 & 15.4 & 78.2 & 10.2 & 88.3 & 7.4 \\
FE2E \cite{wang2025editor} & VAE & 21.9 & 67.2 & 24.1 & 59.6 & 12.0 & 84.0 & 15.1 & 78.5 & 10.0 & 88.6 & 7.1 \\
Diffusion-E2E-FT$^*$ \cite{garcia2025fine} & VAE & 31.7 & 50.9 & 23.3 & \underline{64.4} & 15.0 & 82.8 & 16.6 & 79.5 & 12.7 & 87.4 & 9.3 \\
\midrule
\multicolumn{13}{c}{\textbf{Robust Depth Estimation Model}} \\
\midrule
DynaDepth \cite{zhang2022towards} & ResNet & 38.1 & 39.4 & 51.2 & 29.4 & 23.9 & 60.6 & 17.2 & 60.8 & 14.4 & \underline{90.1} & 13.8 \\
EC-Depth \cite{song2023ec} & ViT-S & 24.3 & 62.3 & 22.8 & 55.2 & 15.5 & 76.6 & 15.8 & 76.7 & 10.9 & 86.1 & 7.6 \\
STEPS \cite{zheng2023steps} & ResNet & 25.2 & 58.8 & 35.0 & 36.7 & 30.1 & 48.0 & 25.2 & 58.8 & 21.6 & 64.1 & 14.8 \\
robustdepth \cite{saunders2023self} & ViT-S & 26.0 & 59.7 & 31.1 & 52.1 & 16.7 & 75.5 & 16.8 & 77.5 & 10.5 & 88.2 & 10.5 \\
weather-depth \cite{wang2024weatherdepth} & ViT-S & - & - & - & - & 15.8 & 76.4 & 16.0 & 76.7 & 10.5 & 87.9 & 12.5 \\
DepthAnything-AC \cite{sun2025depth} & ViT-S & \underline{19.8} & \underline{72.7} & \underline{22.7} & 55.5 & 12.5 & 84.0 & \underline{14.9} & \underline{80.1} & 10.3 & 88.9 & \underline{3.2} \\
\midrule
\textbf{OpenDepth (Ours)} & VAE & \textbf{14.5} & \textbf{81.2} & \textbf{17.0} & \textbf{73.3} & \textbf{9.7} & \textbf{89.3} & \textbf{12.5} & \textbf{84.2} & \textbf{8.5} & \textbf{92.0} & \textbf{1.0} \\
\bottomrule
\end{tabular}
}
\end{table*}

\section{Experiments}

\subsection{Experiment Setup}
\textbf{Implementation details.} We build our method upon the Step1X-Edit \cite{liu2025step1x} v1.0 framework. During finetuning, all parameters except for the DiT module are frozen, and the language control input is left blank. The process employs LoRA with rank = 64 and scale factor $\alpha = 32$. We trained for 20 epochs using the AdamW optimizer with an initial learning rate of $1 \times 10^{-4}$ and a batch size of 8. With gradient checkpointing enabled, the entire training was conducted on NVIDIA RTX 4090 GPUs.

\textbf{Training datasets.} We train our model on a mixture of three datasets: Hypersim \cite{roberts2021hypersim}, Virtual KITTI \cite{cabon2020virtual}, and Cityscapes \cite{cordts2016cityscapes}. Following the data split and preprocessing protocol of Lotus \cite{he2024lotus}, we use approximately 39k images from Hypersim (a photorealistic indoor synthetic dataset), 20k samples from Virtual KITTI (a synthetic street-view dataset with a maximum depth of 80m), and additionally incorporate 20k samples from Cityscapes, a real-world urban driving dataset. Emulating Marigold's \cite{ke2024repurposing} dataset probability sampling mechanism, we adopt a per-dataset sampling ratio of 8:1:1 for Hypersim, Virtual KITTI, and Cityscapes, respectively. 

\textbf{Perturbation strategies.} To synthesize the perturbed counterparts $x_p$, we utilize visual degradations (e.g., low-light, fog, snow, blur) following DepthAnything-AC \cite{sun2025depth}. Unlike applying single random perturbations, we adopt a rigorous compounding strategy during training: low-light degradation with noise is consistently applied as a base perturbation to all $x_p$ samples, while other environmental degradations are probabilistically injected on top in a mutually exclusive manner. This forces the model to learn resilient representations under multi-factor interference. Detailed perturbation configurations are provided in the Appendix.

\textbf{Evaluation benchmarks.} We evaluate the zero-shot performance of our method on both adverse-condition and standard benchmarks following DepthAnything-AC \cite{sun2025depth}. Specifically, we use NuScenes-night \cite{caesar2020nuscenes}, RobotCar-night \cite{maddern20171}, and Driving-Stereo \cite{yang2019drivingstereo} for real-world adverse conditions, and KITTI-C \cite{kong2023robodepth} for synthetic corruptions. For standard generalization, we report results on KITTI \cite{geiger2012we}, NYUv2 \cite{silberman2012indoor}, Sintel \cite{butler2012naturalistic}, ETH3D \cite{schops2017multi}, and DIODE \cite{vasiljevic2019diode}. For all relative-depth benchmarks, predictions are resized to the ground-truth resolution, aligned per image in disparity space on the cropped valid region, clipped to the dataset-specific depth range, and evaluated with AbsRel and $\delta_1$, averaged over images.

\begin{figure*}[t]
    \centering
    \includegraphics[width=0.85\textwidth]{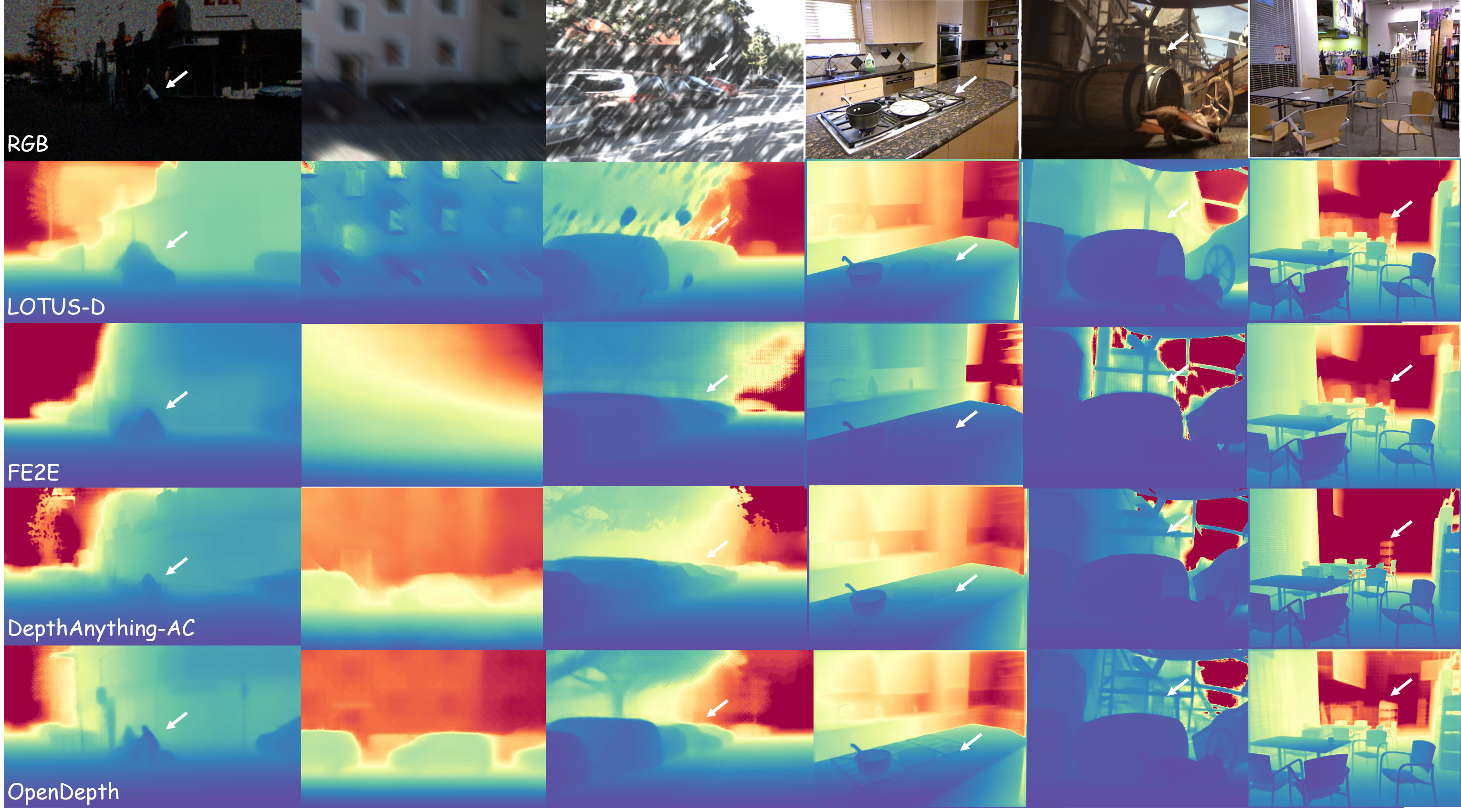}
   \caption{\textbf{Qualitative comparison on zero-shot depth estimation.} OpenDepth demonstrates superior detail preservation and structural consistency over Lotus-D \cite{he2024lotus}, FE2E \cite{wang2025editor}, and DepthAnything-AC \cite{sun2025depth} across both general and open-world conditions, as highlighted by white arrows.}
    \label{fig:comparison1}
\end{figure*}

\subsection{Zero-Shot Relative Depth Estimation on General Benchmarks} 
\textbf{Performance on general benchmarks.} As demonstrated by the average rank in Table~\ref{tab:general_benchmarks}, our OpenDepth model achieves SOTA performance with a top rank of 1.1, successfully outperforming both the general depth estimation models and the specialized robust depth estimation models. Furthermore, the results indicate that I2I-based models, specifically FE2E \cite{wang2025editor} and ours, consistently surpass Text-to-Image models built on pre-trained Stable Diffusion. This clearly validates the superiority of editing priors over generative priors even in standard, general scenarios. Finally, while many foundation models rely on massive amounts of unlabeled training data to achieve strong zero-shot generalization, our method achieves leading performance using merely 79K labeled samples. This highlights the exceptional training data efficiency of our proposed framework.

\subsection{Zero-Shot Relative Depth Estimation on Open-World Benchmarks} 
\textbf{Performance on real complex lighting and weather scenarios.} OpenDepth achieves SOTA results across all open-world adverse benchmarks (e.g., NuScenes-night, DS-rain, as detailed in Table~\ref{tab:open_world}), significantly outperforming both general and specialized robust models. This proves our architecture effectively leverages editing priors to maintain stable geometry perception under severe visual degradation (see visual comparisons in Figure~\ref{fig:comparison1}).

\begin{table*}[t]
\centering
\footnotesize 
\caption{\textbf{Zero-shot relative depth estimation results on synthetic KITTI-C \cite{kong2023robodepth} benchmarks.}}
\label{tab:kitti_c}
\setlength{\tabcolsep}{3.8pt} 
\renewcommand{\arraystretch}{1.1}
\begin{tabular}{l c | c c | c c | c c | c c}
\toprule
\multirow{2}{*}{\textbf{Method}} & \multirow{2}{*}{\textbf{Encoder}} & \multicolumn{2}{c|}{\textbf{Dark}} & \multicolumn{2}{c|}{\textbf{Snow}} & \multicolumn{2}{c|}{\textbf{Motion}} & \multicolumn{2}{c}{\textbf{Gaussian}} \\
\cmidrule(lr){3-4} \cmidrule(lr){5-6} \cmidrule(lr){7-8} \cmidrule(l){9-10}
& & AbsRel$\downarrow$ & $\delta_1\uparrow$ & AbsRel$\downarrow$ & $\delta_1\uparrow$ & AbsRel$\downarrow$ & $\delta_1\uparrow$ & AbsRel$\downarrow$ & $\delta_1\uparrow$ \\
\midrule
DynaDepth \cite{zhang2022towards} & ResNet & 16.3 & 75.2 & 33.8 & 39.3 & 23.4 & 60.9 & 27.4 & 50.1 \\
STEPS \cite{zheng2023steps} & ResNet & 23.0 & 63.1 & 24.2 & 62.2 & 29.1 & 50.8 & 20.4 & 69.2 \\
DepthPro \cite{bochkovskii2024depth} & ViT-S & 14.5 & 79.3 & 19.7 & 68.5 & 17.0 & 74.6 & 17.0 & 74.5 \\
DepthAnything V2 \cite{yang2024depthv2} & ViT-S & \underline{13.0} & 83.2 & \underline{11.5} & \underline{87.2} & 12.7 & 84.0 & 15.7 & 78.5 \\
DepthAnything-AC \cite{sun2025depth} & ViT-S & \underline{13.0} & \underline{83.4} & \textbf{11.4} & \textbf{87.3} & \underline{12.6} & \underline{84.1} & \underline{15.3} & \underline{79.3} \\
\midrule
\textbf{OpenDepth (Ours)} & VAE & \textbf{12.6} & \textbf{83.9} & 11.7 & 86.9 & \textbf{11.8} & \textbf{85.3} & \textbf{14.1} & \textbf{81.3} \\
\bottomrule
\end{tabular}
\end{table*}

\textbf{Performance on KITTI-C benchmarks.} As shown in Table~\ref{tab:kitti_c}, OpenDepth demonstrates exceptional resilience against synthetic corruptions, securing the best performance in three out of four categories (Dark, Motion, and Gaussian noise). This confirms our mechanisms successfully filter out severe appearance shifts for consistent depth prediction.

\begin{table*}[!htbp]
\centering
\footnotesize 
\caption{\textbf{Ablation study of different components.} T2I (Lotus-D \cite{he2024lotus}) and Ours(G) are included as baselines for comparison. (G) and (D) denote generative and discriminative models. $P$: perturbation, $L_{feat}$: feature alignment, $L_{recon}$: RGB reconstruction.}
\label{tab:ablation}
\setlength{\tabcolsep}{2.5pt} 
\renewcommand{\arraystretch}{1.0}
\begin{tabular}{l | c c c | c c | c c | c c | c c}
\toprule
\multirow{2}{*}{\textbf{Baseline}} & \multicolumn{3}{c|}{\textbf{Modules}} & \multicolumn{2}{c|}{\textbf{DS-Rain \cite{yang2019drivingstereo}}} & \multicolumn{2}{c|}{\textbf{DS-Fog \cite{yang2019drivingstereo}}} & \multicolumn{2}{c|}{\textbf{NuScenes-night \cite{caesar2020nuscenes}}} & \multicolumn{2}{c}{\textbf{RobotCar-night \cite{maddern20171}}} \\ 
\cmidrule(lr){2-4} \cmidrule(lr){5-6} \cmidrule(lr){7-8} \cmidrule(lr){9-10} \cmidrule(l){11-12}
 & $P$ & $L_{feat}$ & $L_{recon}$ & AbsRel$\downarrow$ & $\delta_1\uparrow$ & AbsRel$\downarrow$ & $\delta_1\uparrow$ & AbsRel$\downarrow$ & $\delta_1\uparrow$ & AbsRel$\downarrow$ & $\delta_1\uparrow$ \\
\midrule
T2I (Lotus-D) \cite{he2024lotus} & & & & 12.2 & 83.8 & 10.2 & 88.3 & 22.1 & 66.8 & 24.4 & 53.4 \\
               & \checkmark & & & 12.0 & 84.6 & 10.0 & 88.9 & 20.1 & 69.8 & 22.5 & 57.3 \\
\midrule
OpenDepth(G)        & & & & 12.1 & 85.1 & 10.2 & 88.8 & 19.3 & 71.9 & 21.2 & 60.4 \\
\midrule
OpenDepth(D)        & & & & 11.9 & 85.5 & 10.3 & 89.4 & 18.0 & 73.7 & 20.4 & 62.5 \\
               & \checkmark & & & 11.8 & 85.9 & 9.7 & 89.7 & 16.6 & 77.0 & 19.6 & 66.1 \\
               & \checkmark & \checkmark & & 11.5 & 86.2 & 9.4 & 90.4 & 14.8 & 79.8 & 18.5 & 71.1 \\
               & \checkmark & \checkmark & \checkmark & \textbf{9.7} & \textbf{89.3} & \textbf{8.5} & \textbf{92.0} & \textbf{14.5} & \textbf{81.2} & \textbf{17.0} & \textbf{73.3} \\
\bottomrule
\end{tabular}
\end{table*}

\begin{figure*}[!t]
    \centering
    \includegraphics[width=0.95\textwidth]{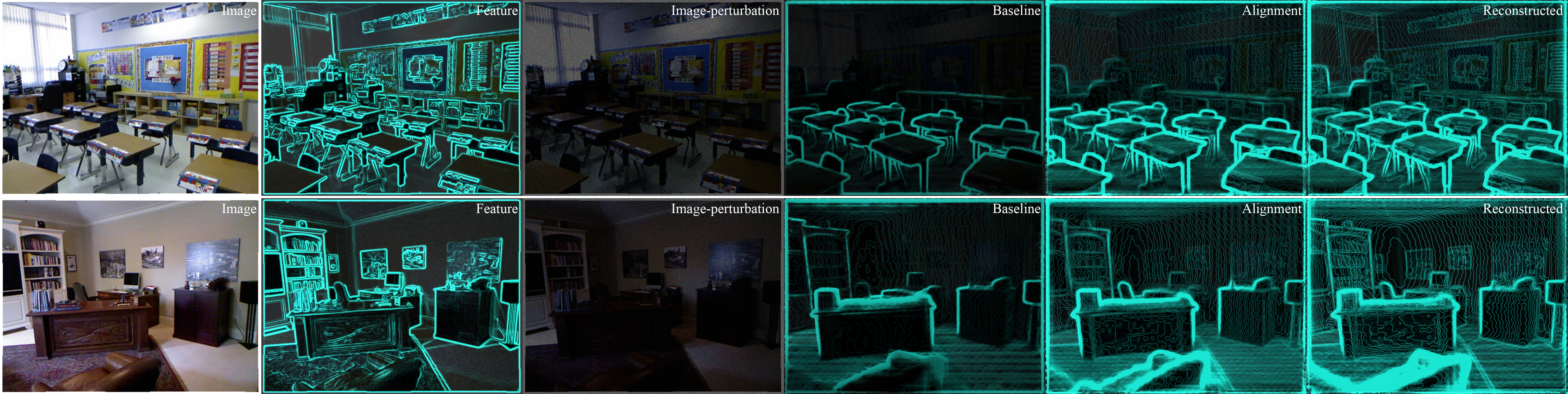}
    \caption{Qualitative comparison of multi-scale edge responses under severe low-light (Dark) perturbations. }
    \label{fig:qualitative_analysis}
\end{figure*}

\subsection{Ablation Study}
To validate the effectiveness of our proposed components, we conduct an ablation study on four open-world adverse condition datasets (see Table~\ref{tab:ablation}). Our baseline employs a DiT-based image editing model whose inherent transformation priors naturally preserve geometric features in complex environments.

We first compare generative (G) and discriminative (D) formulations of our OpenDepth baseline. As shown in Table~\ref{tab:ablation}, OpenDepth(D) consistently outperforms OpenDepth(G) across all benchmarks (e.g., AbsRel 18.0 vs. 19.3 on NuScenes-night), corroborating the finding in Lotus \cite{he2024lotus} that deterministic prediction is more suitable than stochastic generation for dense geometric tasks. Consequently, all OpenDepth results reported in the main comparison tables refer to this discriminative formulation.

\textbf{Effects of Perturbation Training ($\mathcal{L}_{depth}(x_p)$).} Introducing perturbation training yields a consistent improvement, reducing the AbsRel error from 18.0 to 16.6 on NuScenes-night. Notably, even applying the same perturbation strategy to the T2I baseline (Lotus-D) only marginally improves its performance, still substantially lagging behind our OpenDepth(D)+$P$ (16.6). This confirms that perturbation training alone cannot compensate for the inherent limitations of T2I priors, and the superiority of our I2I foundation.

\textbf{Effects of Feature Alignment ($\mathcal{L}_{feat}$).} Adding $\mathcal{L}_{feat}$ further reduces AbsRel from 16.6 to 14.8 on NuScenes-night and from 19.6 to 18.5 on RobotCar-night, demonstrating that our alignment mechanism effectively anchors perturbed inputs to the same fine-grained geometric representations as clean images.

\textbf{Effects of Unified RGB Reconstruction ($\mathcal{L}_{recon}$).} Integrating $\mathcal{L}_{recon}$ yields the most pronounced gains, with AbsRel decreasing from 9.4 to 8.5 on DS-fog and from 18.5 to 17.0 on RobotCar-night. This confirms its critical role in projecting diverse degraded observations into a unified latent domain, realizing robust ``many-to-one'' geometric translation in open-world scenarios.

\subsection{Qualitative Analysis}
As shown in Figure~\ref{fig:comparison1}, , we visualize multi-scale edge responses as geometric proxies under severe low-light (Dark) perturbations (Figure~\ref{fig:qualitative_analysis}). Specifically, these responses are extracted by fusing Sobel gradients, Canny edges, and Laplacian responses from predicted depth maps. The \textbf{Baseline} suffers from severe structural degradation with blurred contours and noisy responses, indicating that distribution shifts can disrupt geometric consistency even with I2I priors. By introducing \textbf{Latent Feature Alignment}, the model suppresses noise and restores structural patterns by anchoring degraded observations to canonical geometric representations. Finally, the \textbf{Full Model (with RGB Reconstruction)} produces sharp and coherent feature maps, accurately recovering fine structures such as chair-leg intersections. This demonstrates that unified RGB reconstruction effectively enforces a style-invariant latent domain for high-fidelity geometric perception.

\section{Conclusion}

In this paper, we propose OpenDepth to elevate robust Monocular Depth Estimation in open-world environments. Through systematic analysis, we demonstrate the inherent superiority of Image-to-Image editing priors over Text-to-Image models in preserving structural integrity against environmental degradations. To resolve the representation misalignment caused by diverse perturbations, we equip our model with ``Intra-Domain Awareness'' via two core mechanisms: shallow feature alignment and joint RGB latent reconstruction. This strategy successfully projects varied degraded inputs into a unified latent space, strictly decoupling stylistic interference from geometric perception. Extensive evaluations confirm that OpenDepth achieves exceptional detail preservation and robust zero-shot generalization, setting new SOTA records across both standard and extreme weather benchmarks.

\section{Acknowledgments}

The authors would like to thank the anonymous reviewers for their constructive comments and suggestions. We also thank all members of our research group for their valuable discussions and support during the development of this work.

\bibliographystyle{ACM-Reference-Format}
\bibliography{references}

\clearpage
\appendix
\twocolumn[
\begin{center}
{\LARGE OpenDepth: Repurposing Image Editing Priors for Robust Open-World Monocular Depth Estimation\par}
\vspace{0.35em}
{\large Supplementary Material\par}
\vspace{1em}
\end{center}
]
\normalsize
\setlength{\tabcolsep}{6pt}
\renewcommand{\arraystretch}{1.0}
\renewcommand{\theHfigure}{supp.\thesection.\arabic{figure}}
\renewcommand{\theHtable}{supp.\thesection.\arabic{table}}
\renewcommand{\theHequation}{supp.\thesection.\arabic{equation}}

\setcounter{section}{0}
\setcounter{subsection}{0}
\setcounter{subsubsection}{0}
\setcounter{figure}{0}
\setcounter{table}{0}
\setcounter{equation}{0}
\renewcommand{\thesection}{\Alph{section}}
\renewcommand{\thesubsection}{\thesection.\arabic{subsection}}
\renewcommand{\thesubsubsection}{\thesubsection.\arabic{subsubsection}}
\renewcommand{\thefigure}{\thesection\arabic{figure}}
\renewcommand{\thetable}{\thesection\arabic{table}}
\renewcommand{\theequation}{\thesection\arabic{equation}}
\makeatletter
\@addtoreset{figure}{section}
\@addtoreset{table}{section}
\@addtoreset{equation}{section}
\makeatother
\noindent
This document provides supplementary information for OpenDepth. We first detail the experimental protocols, datasets, and robust training pipeline. We then present extensive ablation studies regarding architectural design and loss objectives. Finally, we provide additional quantitative benchmarks on inference efficiency and expanded qualitative visualizations to further validate the robustness and generalization of our method.

\section{Experimental Settings}
\label{app:experimental_settings}

\subsection{Evaluation Datasets}
\label{app:test_datasets}
\noindent
To comprehensively assess both robustness in challenging environments and generalization to standard scenes, we evaluate the proposed model on three groups of benchmarks: real-world complex benchmarks, a synthetic corruption benchmark, and general-purpose depth estimation benchmarks.

\noindent\textbf{Real-world complex benchmarks.}
To measure performance under real-world adverse conditions, we adopt NuScenes-night \cite{caesar2020nuscenes}, RobotCar-night \cite{maddern20171}, and DrivingStereo \cite{yang2019drivingstereo}. NuScenes-night is a nighttime subset of the NuScenes benchmark \cite{caesar2020nuscenes} and provides realistic autonomous-driving scenes captured under low-illumination conditions. Following the standard split, we evaluate on 500 nighttime images. RobotCar-night is derived from the Oxford RobotCar benchmark \cite{maddern20171} and contains urban driving scenes captured at night; we use 186 images for evaluation. DrivingStereo \cite{yang2019drivingstereo} is a large-scale real-world driving dataset collected under diverse weather conditions. We use its foggy, rainy, and cloudy subsets, each containing 500 images, to assess robustness under degraded visibility and adverse weather.

\noindent\textbf{Synthetic complex benchmark.}
We further evaluate robustness on KITTI-C \cite{kong2023robodepth}, a corruption benchmark constructed from KITTI \cite{geiger2012we} by introducing controlled synthetic degradations. KITTI-C simulates a range of challenging conditions commonly encountered by visual perception systems. In our experiments, we consider four representative corruption types: dark, snow, motion blur, and Gaussian noise. These settings cover low illumination, severe weather, dynamic blur, and sensor noise, respectively, and provide a systematic testbed for robustness evaluation.

\noindent\textbf{General benchmarks.}
To verify that the robustness enhancement does not compromise performance on standard scenes, we additionally evaluate on KITTI \cite{geiger2012we}, NYUv2 \cite{silberman2012indoor}, Sintel \cite{butler2012naturalistic}, ETH3D \cite{schops2017multi}, and DIODE \cite{vasiljevic2019diode}. KITTI \cite{geiger2012we} is a canonical benchmark for outdoor driving scenes, and we follow the standard split with 652 test images. NYUv2 \cite{silberman2012indoor} is a standard indoor RGB-D benchmark containing diverse household and office environments, and we evaluate on 654 test images. Sintel \cite{butler2012naturalistic} is a synthetic benchmark with complex motion, atmospheric effects, and significant appearance variations; we use the Final split and evaluate on 1,064 images. ETH3D \cite{schops2017multi} provides high-resolution indoor and outdoor scenes with accurate depth annotations, and we evaluate on 454 images. DIODE \cite{vasiljevic2019diode} is a high-quality RGB-D benchmark covering a wide range of scenes and depth scales; following common practice, we report results on its indoor validation split, which contains 325 images.

\noindent\textbf{Evaluation Protocol for Fair Comparison.}
To ensure a strictly fair head-to-head comparison and avoid discrepancies caused by different testing pipelines, we did not simply copy numbers from the original papers. Instead, all baseline results reported in our tables were entirely re-evaluated using their official pre-trained weights. Specifically, for the robust monocular depth estimation baselines, we strictly followed the evaluation settings and codebase provided by DepthAnything-AC \cite{sun2025depth} to reproduce their results. For all other general depth estimation baselines, we re-evaluated them under the exact same dataset-specific evaluation crops, valid depth ranges, and alignment strategies. Furthermore, no extra post-processing steps (e.g., test-time augmentation or multi-scale ensembling) were applied to any of the evaluated methods. This rigorous standardization guarantees that our comparative analysis reflects the true algorithmic capabilities of each method.

\subsection{Evaluation Metrics}
\label{app:evaluation_metrics}

We compute all metrics on the valid pixel set $\Omega$ after applying the dataset-specific crop, validity mask, and evaluation depth range. We report Absolute Relative Error (AbsRel) and threshold accuracy at $1.25$, denoted as $\delta_1$ (D1 in our tables), defined as
\begin{equation}
\mathrm{AbsRel} = \frac{1}{|\Omega|}\sum_{i \in \Omega} \frac{|\hat{d}_i - d_i|}{d_i},
\end{equation}
\begin{equation}
\delta_1 = \frac{1}{|\Omega|}\sum_{i \in \Omega} \mathbf{1}\!\left(
\max\!\left(\frac{\hat{d}_i}{d_i}, \frac{d_i}{\hat{d}_i}\right) < 1.25
\right),
\end{equation}
where $\hat{d}_i$ and $d_i$ denote the predicted and ground-truth depth at pixel $i$, respectively.

Before computing the metrics, we align predictions to the ground truth on the same evaluation region using scale-and-shift fitting in disparity space. Given the raw prediction $p_i$, we solve
\begin{equation}
(a^*, b^*) = \arg\min_{a,b} \sum_{i \in \Omega}
\left(a p_i + b - \frac{1}{d_i}\right)^2,
\end{equation}
and convert the aligned prediction back to depth as
\begin{equation}
\hat{d}_i = \frac{1}{\max(a^* p_i + b^*, \epsilon)}.
\end{equation}
This alignment removes global scale ambiguity while preserving the relative depth structure.

\section{Training Details for the Finetuning Analysis}

\subsection{Perturbations}
\label{app:perturbation}

To enhance the robustness of our depth estimation model, we follow the perturbation strategy of DepthAnything-AC \cite{sun2025depth}. Table~\ref{tab:perturbations} summarizes the categories, probabilities, severity ranges, and detailed generation processes of the perturbations applied during training.

\begin{table*}[h]
  \centering
  \caption{Summary of image perturbations applied during training. Mutually exclusive groups within a category are denoted by matching symbols ($^*$ for weather, $^\dagger$ for blur).}
  \label{tab:perturbations}
  \renewcommand{\arraystretch}{1.4}
  \begin{tabular}{llcc p{8.5cm}}
    \toprule
    \multicolumn{1}{c}{\textbf{Category}} & \textbf{Perturbation} & \textbf{Prob.} & \textbf{Sev.} & \textbf{Process} \\
    \midrule
    \multicolumn{1}{c}{\textbf{Light}} 
    & Dark & 100\% & $1 \sim 5$ & Brightness is decreased nonlinearly ($\gamma=2$), with brighter regions decreasing more and darker regions less. Poisson noise is added to model random photon noise due to reduced photon counts in low-light environments, additional sensor read noise is added to model the read noise of the camera sensor. \\
    \midrule
    \multicolumn{1}{c}{\textbf{Color}} 
    & Contrast & 20\% & $1 \sim 3$ & A centered scaling transform is applied to pixel intensities, reducing the overall image contrast using severity-dependent coefficients. \\
    \midrule
    \multicolumn{1}{c}{\multirow{2}{*}{\textbf{Weather}}} 
    & Fog$^*$ & 20\% & $1 \sim 3$ & A 2D self-similar noise field is generated using the Diamond-Square algorithm, where the severity parameter controls the fog density and texture amplitude, leading to reduced detail retention. \\
    \cmidrule{2-5}
    & Snow$^*$ & 20\% & $1 \sim 2$ & A Gaussian random field is generated as the snow distribution, with snowflake size and density adjusted via nonlinear transformations. Directional motion blur ($\theta \in [-135^\circ, -45^\circ]$) is applied to simulate falling snow. \\
    \midrule
    \multicolumn{1}{c}{\multirow{2}{*}{\textbf{Blur}}} 
    & Motion Blur$^\dagger$ & 10\% & $1 \sim 3$ & A directional blur kernel is constructed via 1D Gaussian sampling with small displacements along a randomly selected motion direction angle $\theta \in [-45^\circ, 45^\circ]$, weighted by Gaussian coefficients. \\
    \cmidrule{2-5}
    & Zoom Blur$^\dagger$ & 10\% & $1 \sim 3$ & Scaling layers with incremental scaling factors are generated using cropping and scaling techniques, and the blurring effect is synthesized by averaging them with equal weights. \\
    \bottomrule
  \end{tabular}
\end{table*}

\subsection{Effect of Training Sampling Probabilities}
\label{app:sampling_probability_effect}

We further analyze the effect of training-set sampling probabilities in mixed-dataset training. Following the current implementation, dataset mixing is realized through \emph{batch-level probabilistic sampling}: at each optimization step, one dataset is selected according to a predefined probability distribution and contributes the current batch. Moreover, since the epoch length is determined by the largest dataloader, changing the sampling ratio directly redistributes the effective optimization budget across datasets rather than merely reweighting samples within a fixed pool. In this study, we denote Hypersim \cite{roberts2021hypersim}, Virtual KITTI \cite{cabon2020virtual}, and Cityscapes \cite{cordts2016cityscapes} by H, V, and C, respectively, and vary their sampling ratios while keeping all other settings unchanged.

\begin{table*}[!htbp]
\centering
\caption{\textbf{Ablation study on training-set sampling probabilities.} All variants are built upon OpenDepth(D) with perturbation training, feature alignment, and RGB reconstruction enabled; only the batch-level dataset sampling ratio is changed. H/V/C denote Hypersim \cite{roberts2021hypersim}, Virtual KITTI \cite{cabon2020virtual}, and Cityscapes \cite{cordts2016cityscapes}, respectively.}
\label{tab:ablation_sampling_probability}
\setlength{\tabcolsep}{3pt} 
\renewcommand{\arraystretch}{1.0}
\begin{tabular}{l | c | c c | c c | c c | c c}
\toprule
\multirow{2}{*}{\textbf{Baseline}} & \multirow{2}{*}{\textbf{Sampling ratio (H:V:C)}} & \multicolumn{2}{c|}{\textbf{DS-rain \cite{yang2019drivingstereo}}} & \multicolumn{2}{c|}{\textbf{DS-fog \cite{yang2019drivingstereo}}} & \multicolumn{2}{c|}{\textbf{NuScenes-night \cite{caesar2020nuscenes}}} & \multicolumn{2}{c}{\textbf{RobotCar-night \cite{maddern20171}}} \\ 
\cmidrule(lr){3-4} \cmidrule(lr){5-6} \cmidrule(lr){7-8} \cmidrule(l){9-10}
 &  & AbsRel$\downarrow$ & $\delta_1\uparrow$ & AbsRel$\downarrow$ & $\delta_1\uparrow$ & AbsRel$\downarrow$ & $\delta_1\uparrow$ & AbsRel$\downarrow$ & $\delta_1\uparrow$ \\
\midrule
OpenDepth(D) & 1:1:1 & 10.4 & 88.4 & 9.1 & 91.0 & 15.5 & 79.6 & 18.0 & 71.8 \\
             & 4:1:1 & 10.0 & 88.8 & 8.7 & 91.6 & 14.9 & 80.7 & 17.5 & 72.6 \\
             & \textbf{8:1:1} & \textbf{9.7} & \textbf{89.3} & \textbf{8.5} & \textbf{92.0} & \textbf{14.5} & \textbf{81.2} & \textbf{17.0} & \textbf{73.3} \\
             & 12:1:1 & 9.9 & 89.0 & 8.6 & 91.8 & 14.8 & 80.9 & 17.4 & 72.8 \\
\midrule
             & 8:2:1 & 9.8 & 89.1 & 8.6 & 91.8 & 14.8 & 80.9 & 17.3 & 72.9 \\
             & 8:1:2 & 10.0 & 88.9 & 8.7 & 91.7 & 14.7 & 81.0 & 17.2 & 73.0 \\
\bottomrule
\end{tabular}
\end{table*}

As shown in Table~\ref{tab:ablation_sampling_probability}, the balanced ratio 1:1:1 yields the weakest overall performance. We attribute this to the reduced sampling frequency of Hypersim \cite{roberts2021hypersim}, which is the most geometry-diverse source in our training mixture and thus provides the strongest structural prior for monocular depth estimation. Increasing the Hypersim proportion from 1:1:1 to 4:1:1 consistently improves all four benchmarks, indicating that a stronger geometry-centered supervision budget is crucial for robust open-world generalization.

When the ratio is further increased to 8:1:1, the model achieves the best trade-off across both adverse weather and nighttime benchmarks. This suggests that the dominant contribution of Hypersim effectively stabilizes generic geometric reasoning, while the smaller fractions of Virtual KITTI \cite{cabon2020virtual} and Cityscapes \cite{cordts2016cityscapes} still provide indispensable outdoor-domain anchors. In particular, Virtual KITTI contributes synthetic driving layouts and weather-relevant scene structure, whereas Cityscapes injects real-world urban appearance statistics that are beneficial for nighttime transfer.

However, pushing the mixture further toward Hypersim (12:1:1) does not yield additional gains and instead causes slight degradation. This implies that, although geometry-rich indoor supervision should dominate the training budget, excessively suppressing outdoor data weakens the model's adaptation to driving-specific scene distributions. Similarly, increasing the share of Virtual KITTI (8:2:1) or Cityscapes (8:1:2) also fails to surpass 8:1:1. The former introduces a stronger synthetic-domain bias, while the latter overemphasizes a comparatively narrow real driving domain, both of which slightly compromise the global balance between geometric diversity and outdoor robustness.

Overall, these results indicate that robust open-world monocular depth estimation benefits from a deliberately \emph{asymmetric} sampling strategy: a geometry-rich source should dominate the optimization process to establish strong structural priors, while smaller yet non-negligible outdoor synthetic and real-world sources are required to regularize domain-specific appearance statistics. This observation justifies our adoption of \textbf{8:1:1} as the default sampling configuration throughout the paper.

\subsection{Full Training Pipeline}
\label{app:full_training_pipeline}

\noindent\textbf{Overview \& Data Preparation.}
Our training framework is built upon Step1X-Edit \cite{liu2025step1x}. During data preparation, the target depth map and the reference RGB image are encoded into VAE latents. In mixed-dataset training, datasets are sampled at every optimization step according to a predefined probability distribution. Since our method does not rely on semantic language control, the text prompt is kept empty, and the text conditions are extracted deterministically via the frozen Qwen2.5-VL \cite{wu2025qwen} encoder.

\noindent\textbf{Generative vs. Discriminative Training.}
For the generative baseline, the model follows the standard flow-matching formulation, where Gaussian noise is added to the target depth latent and a timestep is drawn from a scheduler. In contrast, our primary \textbf{discriminative} variant replaces stochastic denoising with direct latent regression. The target depth latent is strictly zero-initialized, the timestep is fixed at $t=1.0$, and the model directly regresses the ground-truth depth latent using a masked Mean Squared Error (MSE) loss over valid pixels.

\noindent\textbf{Robust Training \& Objectives.}
Our robust training pipeline extends the discriminative formulation by jointly optimizing a clean branch and a perturbed branch. For each batch, both the clean image and its perturbed counterpart are processed. The perturbed branch is supervised using the exact same ground-truth depth latent as the clean branch to enforce output-level invariance. Furthermore, the overall robust objective integrates shallow feature alignment (cosine consistency between perturbed features and stopped-gradient clean features) and unified RGB reconstruction (mapping the perturbed RGB output toward the clean RGB latent). 

\noindent\textbf{Optimization Details.}
To ensure training efficiency, we freeze the base Step1X-Edit \cite{liu2025step1x} DiT backbone, the VAE, and the text encoder, introducing trainable parameters exclusively through LoRA adapters. The training process leverages standard memory-efficient techniques, including mixed precision, gradient checkpointing, and fp8 base weights.

\section{Ablation on Design Details}
\label{app:ablation_details}

\subsection{Sensitivity to Loss Weights}
\label{app:ablation_loss_sensitivity}

We further analyze the sensitivity of the final objective
\[
\mathcal{L}_{total} = \mathcal{L}_{depth}(x_c) + \alpha \mathcal{L}_{depth}(x_p) + \beta \mathcal{L}_{feat} + \gamma \mathcal{L}_{recon},
\]
where $\mathcal{L}_{recon}$ denotes the cross-image reconstruction term only. We vary one coefficient at a time while fixing the other two to $1$, and evaluate all variants on the same four robustness benchmarks used in the main paper.

\begin{table*}[!htbp]
\centering
\caption{\textbf{Ablation study on loss weights.} All variants are built upon OpenDepth(D) with perturbation training, feature alignment, and cross-image reconstruction enabled; only the loss weights in the final objective are changed.}
\label{tab:ablation_loss_weights}
\setlength{\tabcolsep}{3pt}
\renewcommand{\arraystretch}{1.0}
\begin{tabular}{l | c | c c | c c | c c | c c}
\toprule
\multirow{2}{*}{\textbf{Baseline}} & \multirow{2}{*}{\textbf{Loss weights ($\alpha$:$\beta$:$\gamma$)}} & \multicolumn{2}{c|}{\textbf{DS-rain \cite{yang2019drivingstereo}}} & \multicolumn{2}{c|}{\textbf{DS-fog \cite{yang2019drivingstereo}}} & \multicolumn{2}{c|}{\textbf{NuScenes-night \cite{caesar2020nuscenes}}} & \multicolumn{2}{c}{\textbf{RobotCar-night \cite{maddern20171}}} \\
\cmidrule(lr){3-4} \cmidrule(lr){5-6} \cmidrule(lr){7-8} \cmidrule(l){9-10}
 &  & AbsRel$\downarrow$ & $\delta_1\uparrow$ & AbsRel$\downarrow$ & $\delta_1\uparrow$ & AbsRel$\downarrow$ & $\delta_1\uparrow$ & AbsRel$\downarrow$ & $\delta_1\uparrow$ \\
\midrule
OpenDepth(D) & 0.5:1:1 & 10.5 & 88.1 & 9.1 & 90.9 & 15.9 & 79.4 & 18.5 & 71.1 \\
             & \textbf{1:1:1} & \textbf{9.7} & \textbf{89.3} & \textbf{8.5} & \textbf{92.0} & \textbf{14.5} & \textbf{81.2} & \textbf{17.0} & \textbf{73.3} \\
             & 2:1:1 & 10.1 & 88.7 & 8.7 & 91.4 & 15.1 & 80.3 & 17.8 & 72.1 \\
\midrule
             & 1:0.5:1 & 10.4 & 88.3 & 8.9 & 91.0 & 15.7 & 79.5 & 18.4 & 71.2 \\
             & 1:2:1 & 10.0 & 88.8 & 8.7 & 91.6 & 15.2 & 80.1 & 17.9 & 72.0 \\
\midrule
             & 1:1:0.5 & 10.3 & 88.4 & 8.9 & 91.1 & 15.4 & 79.9 & 18.2 & 71.5 \\
             & 1:1:2 & 9.9 & 88.9 & 8.6 & 91.7 & 14.9 & 80.6 & 17.6 & 72.4 \\
\bottomrule
\end{tabular}
\end{table*}

Table~\ref{tab:ablation_loss_weights} shows that the balanced setting $\alpha:\beta:\gamma=1:1:1$ gives the best overall trade-off across all four benchmarks. Decreasing $\alpha$ leads to the clearest performance drop, especially on the two night benchmarks, indicating that explicit supervision on perturbed inputs is essential for robust generalization. Increasing $\alpha$ to $2$ partially preserves robustness, but still underperforms the balanced setting, suggesting that overly emphasizing noisy-view fitting can bias optimization away from stable geometric learning. Varying $\beta$ reveals that feature alignment is particularly important for cross-domain nighttime robustness: a small $\beta$ is insufficient to stabilize the latent representation, while a large $\beta$ slightly over-regularizes intermediate features. Finally, $\gamma$ mainly controls appearance-invariant regularization through cross-image reconstruction. Reducing it weakens robustness under adverse appearance shifts, while increasing it improves robustness only marginally and remains inferior to the balanced choice. Overall, the three auxiliary terms are complementary, and we therefore adopt $\alpha=\beta=\gamma=1$ in all experiments.

\subsection{Analysis of the Perturbation Branch Supervision}
\label{app:ablation_branch_supervision}

During robustness training, the perturbed-input branch requires a depth supervision target.
Two natural candidates exist:
(i)~\textbf{Fit GT}, which directly regresses the ground-truth depth in the VAE latent space
($\mathcal{L}_\text{pert}=\lVert\hat{d}_\text{pert}-d_\text{gt}\rVert^2$); and
(ii)~\textbf{Fit Clean}, a self-distillation strategy that regresses the stopped-gradient prediction
of the clean-input branch
($\mathcal{L}_\text{pert}=\lVert\hat{d}_\text{pert}-\mathrm{sg}(\hat{d}_\text{clean})\rVert^2$).
The two variants share the same backbone, perturbation strategy, feature alignment,
and RGB reconstruction; only the depth target differs.

\begin{table*}[!htbp]
\centering
\caption{\textbf{Ablation on the supervision target for the perturbed branch.}
``Fit GT'' regresses ground-truth depth; ``Fit Clean'' regresses the stopped-gradient clean-branch prediction.}
\label{tab:ablation_target}
\setlength{\tabcolsep}{3pt}
\renewcommand{\arraystretch}{1.05}
\begin{tabular}{l | c c | c c | c c | c c}
\toprule
\multirow{2}{*}{\textbf{Target}}
  & \multicolumn{2}{c|}{\textbf{DS-rain \cite{yang2019drivingstereo}}}
  & \multicolumn{2}{c|}{\textbf{DS-fog \cite{yang2019drivingstereo}}}
  & \multicolumn{2}{c|}{\textbf{NuScenes-night \cite{caesar2020nuscenes}}}
  & \multicolumn{2}{c}{\textbf{RobotCar-night \cite{maddern20171}}} \\
\cmidrule(lr){2-3}\cmidrule(lr){4-5}\cmidrule(lr){6-7}\cmidrule(l){8-9}
  & AbsRel$\downarrow$ & $\delta_1\uparrow$
  & AbsRel$\downarrow$ & $\delta_1\uparrow$
  & AbsRel$\downarrow$ & $\delta_1\uparrow$
  & AbsRel$\downarrow$ & $\delta_1\uparrow$ \\
\midrule
Fit Clean
  & 10.3 & 88.5
  & 9.0  & 91.1
  & 15.3 & 79.8
  & 17.8 & 71.9 \\
\textbf{Fit GT (ours)}
  & \textbf{9.7}  & \textbf{89.3}
  & \textbf{8.5}  & \textbf{92.0}
  & \textbf{14.5} & \textbf{81.2}
  & \textbf{17.0} & \textbf{73.3} \\
\bottomrule
\end{tabular}
\end{table*}

As shown in Table~\ref{tab:ablation_target}, Fit~GT consistently outperforms Fit~Clean
on all four adverse-condition benchmarks,
reducing AbsRel by 0.5--0.8 and raising $\delta_1$ by 0.8--1.4\,pp.
The gap is most pronounced on the two nighttime sets
(NuScenes-night and RobotCar-night),
where AbsRel drops by 0.8 in both cases.
This is expected: nighttime scenes lie far outside the clean training distribution,
so the clean branch itself produces noisy predictions;
using them as pseudo-labels introduces a \emph{confirmation bias}
that caps the perturbed branch's potential.
Ground-truth supervision, by contrast, provides an oracle signal
invariant to input corruptions, yielding lower-variance gradients
and a strictly higher accuracy ceiling.
We therefore adopt Fit~GT as the default target in all experiments.

\subsection{Sensitivity to the Number of Feature Alignment Layers}
\label{app:ablation_feature_layers}

\begin{table*}[!htbp]
\centering

\caption{\textbf{Ablation study on the depth and position of feature alignment.} All variants are built upon OpenDepth(D) with perturbation training, feature alignment, and RGB reconstruction enabled; only the aligned Single Stream layers are changed. First N denotes aligning the first $N$ Single Stream blocks. Middle 4 and Last 4 denote aligning the middle four and last four Single Stream blocks, respectively.}
\label{tab:ablation_feat_layers}
\setlength{\tabcolsep}{3pt} 
\renewcommand{\arraystretch}{1.0}
\begin{tabular}{l | c | c c | c c | c c | c c}
\toprule
\multirow{2}{*}{\textbf{Baseline}} & \multirow{2}{*}{\textbf{Aligned layers}} & \multicolumn{2}{c|}{\textbf{DS-rain \cite{yang2019drivingstereo}}} & \multicolumn{2}{c|}{\textbf{DS-fog \cite{yang2019drivingstereo}}} & \multicolumn{2}{c|}{\textbf{NuScenes-night \cite{caesar2020nuscenes}}} & \multicolumn{2}{c}{\textbf{RobotCar-night \cite{maddern20171}}} \\ 
\cmidrule(lr){3-4} \cmidrule(lr){5-6} \cmidrule(lr){7-8} \cmidrule(l){9-10}
 &  & AbsRel$\downarrow$ & $\delta_1\uparrow$ & AbsRel$\downarrow$ & $\delta_1\uparrow$ & AbsRel$\downarrow$ & $\delta_1\uparrow$ & AbsRel$\downarrow$ & $\delta_1\uparrow$ \\
\midrule
OpenDepth(D) & First 1 & 10.0 & 88.8 & 8.8 & 91.5 & 14.9 & 80.5 & 17.7 & 72.3 \\
             & First 2 & 9.8 & 89.1 & 8.6 & 91.8 & 14.7 & 80.9 & 17.4 & 72.9 \\
             & \textbf{First 4} & \textbf{9.7} & \textbf{89.3} & \textbf{8.5} & \textbf{92.0} & \textbf{14.5} & \textbf{81.2} & \textbf{17.0} & \textbf{73.3} \\
             & First 8 & 9.9 & 89.0 & 8.6 & 91.8 & 14.8 & 80.7 & 17.3 & 72.8 \\
\midrule
             & Middle 4 & 10.0 & 88.8 & 8.7 & 91.6 & 15.0 & 80.2 & 17.6 & 72.0 \\
             & Last 4 & 10.2 & 88.5 & 8.9 & 91.2 & 15.3 & 79.5 & 17.9 & 71.3 \\
\bottomrule
\end{tabular}
\end{table*}
\noindent\textbf{Effect of alignment depth.}
Table~\ref{tab:ablation_feat_layers} shows a clear trend with respect to the number of aligned shallow Single Stream blocks. Increasing the aligned prefix from the first layer to the first four layers consistently improves all benchmarks. Compared with aligning only the first layer, aligning the first four layers reduces AbsRel from 10.0 to 9.7 on DS-rain, from 8.8 to 8.5 on DS-fog, from 14.9 to 14.5 on NuScenes-night, and from 17.7 to 17.0 on RobotCar-night, while improving $\delta_1$ by 0.5, 0.5, 0.7, and 1.0 points, respectively. This result indicates that aligning only one or two layers is insufficient to fully suppress perturbation-induced feature drift once the fused representation propagates deeper into the network. Although the first two layers already provide noticeable gains, the first four layers achieve a more favorable trade-off by stabilizing a larger portion of the shallow fused representation.

However, extending the aligned region further to the first eight Single Stream blocks does not bring additional benefits and instead causes slight but consistent degradation on all datasets. For example, compared with \emph{First 4}, \emph{First 8} increases AbsRel from 14.5 to 14.8 on NuScenes-night and from 17.0 to 17.3 on RobotCar-night, while also reducing $\delta_1$. This suggests that deeper alignment begins to over-constrain mid-level fused features, limiting the model's flexibility to adapt its internal representation for downstream geometric reasoning.

\noindent\textbf{Effect of alignment position.}
When the number of aligned layers is fixed to four, the position of alignment becomes equally important. Aligning the first four Single Stream blocks clearly outperforms aligning the middle four or the last four. Relative to \emph{Middle 4}, \emph{First 4} achieves lower AbsRel by 0.3 on DS-rain, 0.2 on DS-fog, 0.5 on NuScenes-night, and 0.6 on RobotCar-night, together with consistent gains in $\delta_1$. The gap becomes even larger when comparing \emph{First 4} against \emph{Last 4}: AbsRel is reduced by 0.5, 0.4, 0.8, and 0.9 on the four benchmarks, respectively. Notably, the largest margins appear on the nighttime benchmarks, where late-layer alignment yields the weakest results.

This behavior is consistent with the functional role of the Single Stream stage. Early Single Stream blocks have already incorporated preliminary cross-modal fusion, yet they remain relatively close to the input structure. They therefore provide an ideal location for anchoring perturbation-invariant geometric representations. In contrast, middle and late Single Stream blocks are increasingly specialized for prediction-oriented reasoning. Applying strong alignment at these stages is less effective for preventing early drift and may interfere with the model's ability to preserve task-specific depth cues under severe domain shift.

\noindent\textbf{Discussion.}
Overall, the ablation reveals a consistent depth-position trade-off for feature alignment: the alignment should be applied sufficiently early to suppress perturbation-induced drift before it propagates through the fused representation, but not so deep that it over-regularizes semantically richer prediction-oriented features. These results substantiate our architectural choice of aligning the first four Single Stream blocks, which provides the most effective balance between perturbation invariance and representational flexibility.

\begin{table*}[t]
  \centering
  \caption{Quantitative comparison of generative and discriminative depth
           estimation models on NYUv2 (cap 10\,m, Eigen crop) and KITTI
           (cap 80\,m, no crop). Alignment: disparity-space least-squares.
           $\downarrow$: lower is better; $\uparrow$: higher is better.
           \textbf{Bold} indicates the better result per metric per dataset.}
  \label{tab:depth_eval}
  \setlength{\tabcolsep}{5.5pt}
  \renewcommand{\arraystretch}{1.15}
  \begin{tabular}{llccccccccc}
    \toprule
    \multirow{2}{*}{\textbf{Model}} & \multirow{2}{*}{\textbf{Dataset}} &
      \multicolumn{6}{c}{\textbf{Error} $\downarrow$} &
      \multicolumn{3}{c}{\textbf{Accuracy (\%)} $\uparrow$} \\
    \cmidrule(lr){3-8}\cmidrule(lr){9-11}
    & & AbsRel (\%) & SqRel & RMSE & RMSElog & SILog & log$_{10}$ &
        $\delta_1$ & $\delta_2$ & $\delta_3$ \\
    \midrule
    \multirow{2}{*}{\textbf{Discriminative}}
      & NYUv2
        &\textbf{3.88}           & \textbf{0.0179} & \textbf{0.1871} & \textbf{0.0599} & \textbf{5.9807} &\textbf{0.0170} 
        & \textbf{98.04} & \textbf{99.42} & \textbf{99.75} \\
      & KITTI
        & \textbf{6.30} & \textbf{0.2924} & \textbf{3.1318} & \textbf{0.0986} & \textbf{9.8265} & \textbf{0.0276}
        & \textbf{96.20} & \textbf{99.22} & \textbf{99.84} \\
    \midrule
    \multirow{2}{*}{\textbf{Generative}}
      & NYUv2
        & 3.90 & 0.0203          & 0.1974          & 0.0616          & 6.1434          &  0.0171
        & 97.85          & 99.35          & 99.74 \\
      & KITTI
        & 6.54          & 0.3885          & 3.3800          & 0.1041          & 10.3756         & 0.0281
        & 95.88          & 99.07          & 99.74 \\
    \bottomrule
  \end{tabular}
\end{table*}

\subsection{Analysis of Generative and Discriminative Models}
\label{app:generative_vs_discriminative}

In this section, we provide a quantitative comparison between the generative and discriminative variants of our depth estimation model. Table~\ref{tab:depth_eval} summarizes the performance of both paradigms on the NYUv2 (indoor) and KITTI (outdoor) datasets. 

Overall, the discriminative model consistently outperforms its generative counterpart across all metrics on both datasets. On the indoor NYUv2 dataset, the discriminative model achieves an AbsRel error of 3.88 and a $\delta_1$ accuracy of 98.04, demonstrating highly precise metric depth recovery. While the generative model yields slightly higher errors (e.g., an AbsRel of 3.90 and an RMSE of 0.1974), the performance gap remains marginal, indicating that the generative paradigm can still synthesize highly accurate indoor depth structures.

Conversely, the performance discrepancy becomes more pronounced on the outdoor KITTI dataset, which features complex, unbounded scenes and large depth ranges. The discriminative model attains an AbsRel of 6.30 and an RMSE of 3.1318, whereas the generative model degrades to an AbsRel of 6.54 and an RMSE of 3.3800. This widening gap suggests that the iterative denoising process inherent to generative models may struggle to maintain strict geometric consistency in unbounded outdoor environments, where scale ambiguity and distant structures pose significant challenges. In contrast, the discriminative model, which directly regresses depth values in a single forward pass, exhibits superior robustness and geometric fidelity across diverse scenarios.

\subsection{Generalization to Unseen Perturbations}
\label{sec:unseen_perturbations}

A critical concern regarding robust training is whether the model merely overfits to the specific degradation patterns seen during training, or if it genuinely learns to extract invariant geometric structures. To investigate this, we evaluate the generalization capability of our model on \textit{unseen} perturbations. Specifically, as detailed in our perturbation strategy (Section~\ref{app:perturbation}), our training pipeline does not include rain, cloudy-weather, or Gaussian noise degradations. 

We evaluate zero-shot performance on DS-rain \cite{yang2019drivingstereo}, DS-cloud \cite{yang2019drivingstereo}, and KITTI-C \cite{kong2023robodepth} Gaussian benchmarks and compare OpenDepth with the state-of-the-art robust baseline, DepthAnything-AC \cite{sun2025depth}. Note that both models are exposed to similar perturbation synthesis strategies during training, making this a fair head-to-head comparison of out-of-distribution generalization.

\begin{table*}[h]
\centering
\caption{\textbf{Generalization to unseen perturbations.} Neither model was exposed to rain, cloud, or Gaussian noise during training.}
\label{tab:unseen_perturbations}
\setlength{\tabcolsep}{4pt} 
\renewcommand{\arraystretch}{1.1}
\begin{tabular}{l | c c | c c | c c}
\toprule
\multirow{2}{*}{\textbf{Method}} & \multicolumn{2}{c|}{\textbf{DS-rain \cite{yang2019drivingstereo}}} & \multicolumn{2}{c|}{\textbf{DS-cloud \cite{yang2019drivingstereo}}} & \multicolumn{2}{c}{\textbf{KITTI-C Gaussian \cite{kong2023robodepth}}} \\
\cmidrule(lr){2-3} \cmidrule(lr){4-5} \cmidrule(l){6-7}
& AbsRel$\downarrow$ & $\delta_1\uparrow$ & AbsRel$\downarrow$ & $\delta_1\uparrow$ & AbsRel$\downarrow$ & $\delta_1\uparrow$ \\
\midrule
DepthAnything-AC \cite{sun2025depth} & 12.5 & 84.0 & 14.9 & 80.1 & 15.3 & 79.3 \\
\textbf{OpenDepth (Ours)} & \textbf{9.7} & \textbf{89.3} & \textbf{12.5} & \textbf{84.2} & \textbf{14.1} & \textbf{81.3} \\
\bottomrule
\end{tabular}
\end{table*}

As shown in Table~\ref{tab:unseen_perturbations}, OpenDepth significantly outperforms DepthAnything-AC \cite{sun2025depth} across all three unseen degradation types. On the DS-rain benchmark, OpenDepth reduces the AbsRel error from 12.5 to 9.7 and improves $\delta_1$ accuracy from 84.0 to 89.3. Substantial improvements are also observed under cloudy conditions (AbsRel: 14.9 $\rightarrow$ 12.5) and severe sensor noise (KITTI-C Gaussian AbsRel: 15.3 $\rightarrow$ 14.1). 

These results strongly confirm that OpenDepth does not simply memorize the training perturbations. Instead, by leveraging the inherent structural preservation capabilities of Image-to-Image editing priors, combined with our proposed latent feature alignment and unified RGB reconstruction, the model successfully learns a generalized ``Intra-Domain Awareness''. This allows it to project entirely novel visual degradations into a canonical clean latent space, demonstrating true out-of-distribution robustness.

\section{Failure Case Analysis}
\label{app:failure_case_analysis}

\begin{figure*}[t]
\centering
\includegraphics[width=\textwidth]{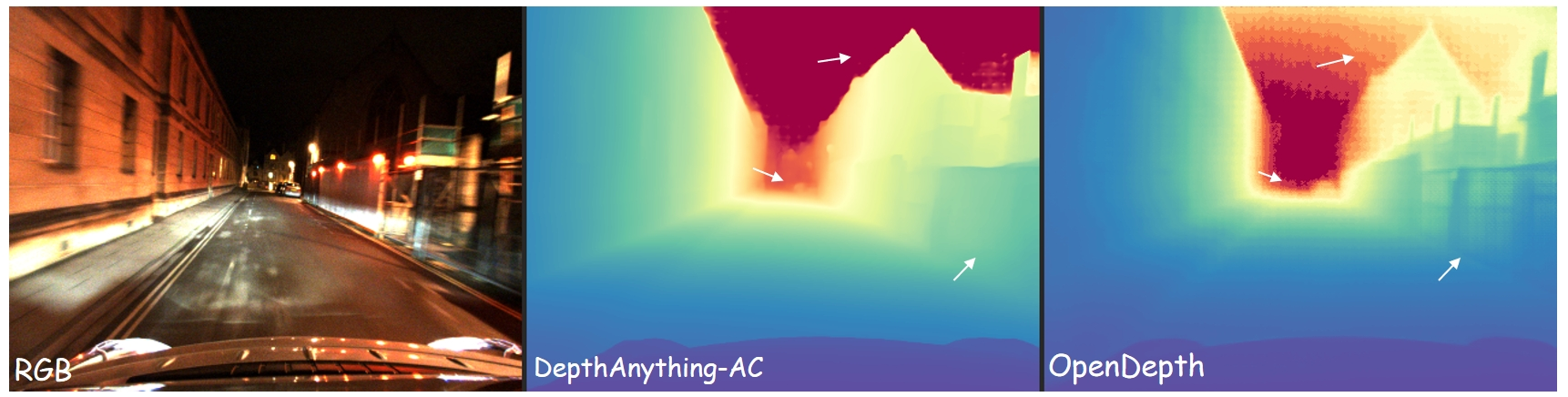}
\caption{\textbf{Representative case with remaining challenges on RobotCar-night \cite{maddern20171} (max distance = 80\,m).}
From left to right: RGB image, DepthAnything-AC \cite{sun2025depth}, and OpenDepth. White arrows highlight that OpenDepth preserves clearer near-range boundaries and more coherent local structure under severe nighttime illumination, while very distant regions remain relatively less separated. In a few extremely dark scenes, sky regions may also be assigned a finite depth rather than the farthest depth.}
\label{fig:failure_cases}
\end{figure*}

Although OpenDepth achieves better overall AbsRel and $\delta_1$ on RobotCar-night \cite{maddern20171}, Fig.~\ref{fig:failure_cases} shows that the remaining errors are not uniformly distributed across depth ranges. As indicated by the white arrows, OpenDepth tends to preserve sharper local geometry in the near field, including building contours, roadside structures, and the geometric grouping of nearby scene components. This tendency is consistent with our robust training pipeline, where clean and perturbed inputs share the same GT depth target, early Single Stream features are aligned to their clean counterparts, and an auxiliary RGB latent reconstruction term encourages recovery of a clean-domain representation from degraded inputs. Together, these objectives appear particularly helpful for stabilizing local structural cues that are easily weakened in low-light driving scenes.

The same example also suggests a characteristic tendency in very distant regions. In the current implementation, the supervision target is not metric depth itself; instead, depth is first converted into an inverse-depth/disparity-style representation, normalized with trunc\_disparity, and then encoded into the VAE latent space for regression. Moreover, evaluation on RobotCar-night \cite{maddern20171} applies scale-and-shift fitting in disparity space under an 80\,m cap. Under this formulation, large metric-depth differences in the far field correspond to relatively small changes in disparity, and the percentile-based truncation further keeps very distant content within a comparatively compact representation range. As a result, the model allocates stronger effective resolution to near- and mid-range geometry, whereas the separation of distant buildings, road extents, and skyline structures can be less pronounced in some nighttime scenes.

A related phenomenon appears in a small number of extremely dark samples, where sky regions are not always mapped to the farthest depth. This behavior is also compatible with the current design: the model regresses a normalized disparity latent rather than an explicit ``infinite-depth'' label, and the auxiliary feature-alignment and RGB-reconstruction losses mainly promote perturbation-invariant representation learning instead of imposing a dedicated sky prior. At the same time, the 80\,m evaluation mask excludes pixels beyond the valid depth range, so this effect usually occupies only a limited portion of the image and has less influence on aggregate metrics than on qualitative inspection. Overall, this case suggests that OpenDepth currently provides a favorable trade-off on nighttime driving benchmarks, offering stronger near-range structural fidelity and robustness while leaving room for further improvement in far-range calibration and sky handling under extremely low illumination.

\begin{table*}[t]
\centering
\caption{Inference efficiency and performance comparison on the whole NYUv2 \cite{silberman2012indoor} dataset (batch size 1) measured on a single NVIDIA GeForce RTX 4090 GPU. All methods are evaluated at the same internal resolution of 768x576.}
\label{tab:inference_efficiency}
\begin{tabular}{lccccccc}
\toprule
Model & Resolution  & Latency (ms) & Throughput (img/s) & Memory (GB) & AbsRel$\downarrow$ & $\delta_1\uparrow$ \\
\midrule
Lotus-D \cite{he2024lotus} & 768x576  & 230.3 & 4.34 & 6.44 & 5.2 & 96.8 \\
FE2E \cite{wang2025editor} & 768x576  & 2714.6 & 0.38 & 25.20 & 4.1 & 97.7 \\
OpenDepth(D) & 768x576 &  704.4 & 1.42 & 25.05 & 3.9 & 98.0 \\
\bottomrule
\end{tabular}
\end{table*}

\section{Inference Efficiency Analysis}
\label{app:inference_efficiency}
\noindent
We benchmark inference efficiency on the whole NYUv2 \cite{silberman2012indoor} dataset with raw resolution 640x480, using batch size 1 on a single NVIDIA GeForce RTX 4090 GPU. To ensure a fair comparison, all methods are evaluated at the same internal resolution of 768x576. 

\textbf{Cached Empty-Prompt Conditioning.} 
To address the computational bottleneck introduced by the heavy large language model (LLM) text encoder, we introduce a cached empty-prompt conditioning strategy during inference. Since our framework does not rely on semantic language control for depth estimation, the text prompt remains strictly empty. Consequently, the output embeddings from the text encoder (i.e., Qwen2.5-VL \cite{wu2025qwen}) are deterministic and constant. By pre-computing and caching these empty-prompt embeddings offline, we completely bypass the need to load and execute the massive Qwen model during actual inference. This strategy effectively ``prunes'' the text encoding branch at runtime, significantly reducing both GPU memory consumption and inference latency without requiring complex architectural distillation or sacrificing any structural priors.

As shown in Table~\ref{tab:inference_efficiency}, with this cached conditioning strategy, Discriminative OpenDepth achieves 704.4 ms latency and 1.42 img/s throughput, making it about 4.0$\times$ faster than FE2E \cite{wang2025editor} while using slightly less memory (25.05 GB vs. 25.20 GB). Lotus-D \cite{he2024lotus} remains the fastest and most lightweight baseline overall due to its much smaller parameter count. However, although our method is slower than Lotus-D \cite{he2024lotus}, it establishes a new SOTA performance over the previous SOTA method, FE2E \cite{wang2025editor}, while simultaneously achieving a 4.0$\times$ speedup in inference.

\section{More Visual Results}
\label{app:more_visual_results}
\noindent

\begin{figure*}[t]
\centering

\includegraphics[width=0.80\textwidth]{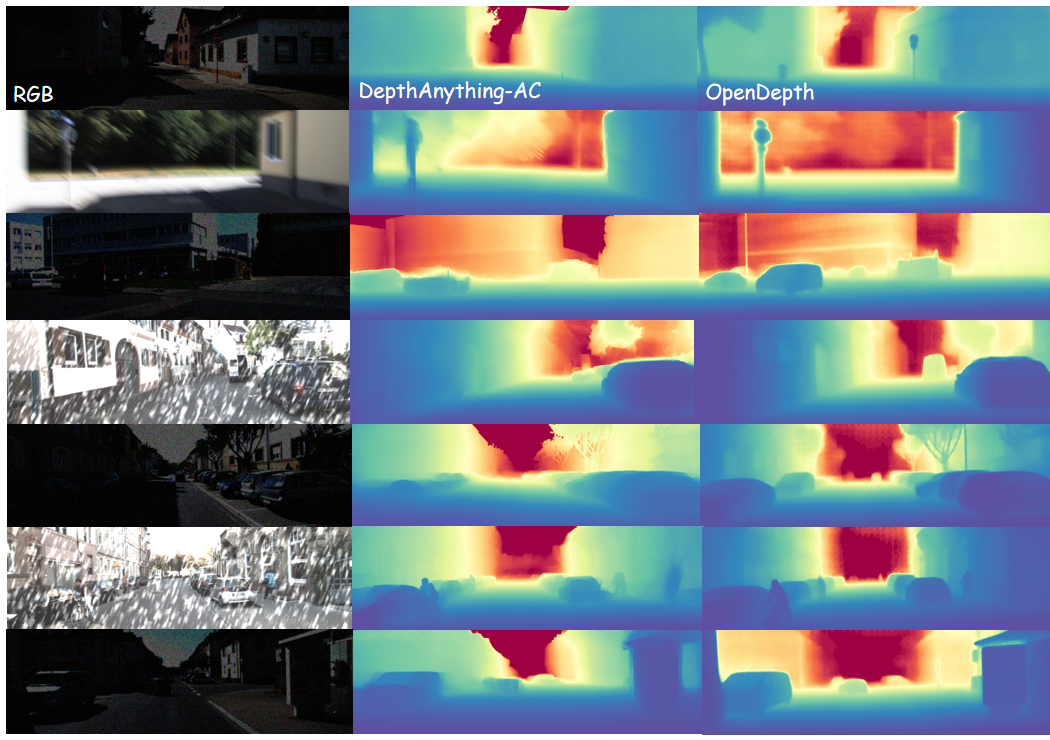}
\caption{Comparison between DepthAnything-AC \cite{sun2025depth} and our OpenDepth}
\label{fig:more1}
\end{figure*}

\begin{figure*}[t]
\centering

\includegraphics[width=0.80\textwidth]{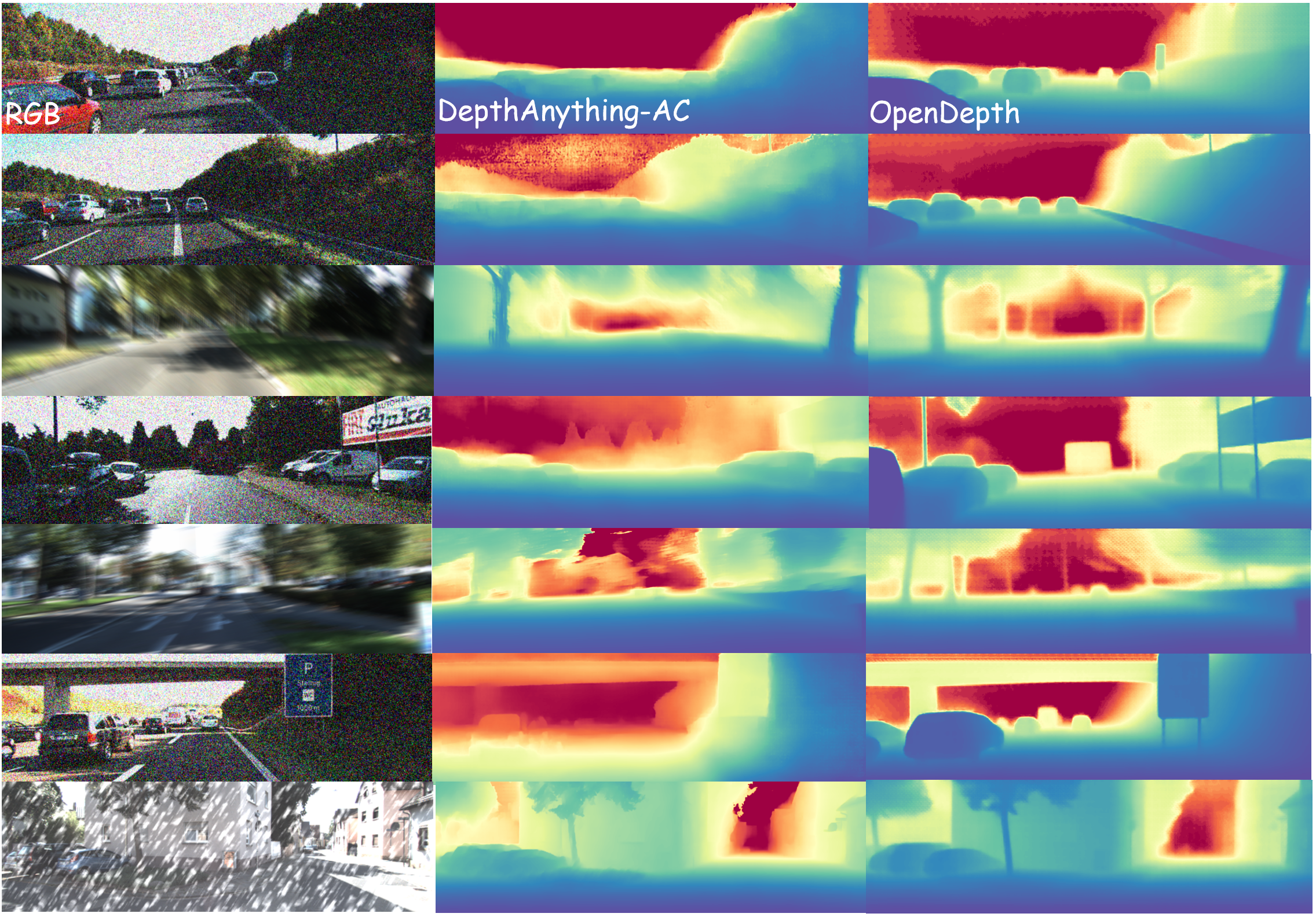}
\caption{Comparison between DepthAnything-AC \cite{sun2025depth} and our OpenDepth}
\label{fig:more2}
\end{figure*}

\end{document}